\documentclass[sigconf,screen, 9pt]{acmart}
\renewcommand\footnotetextcopyrightpermission[1]{} 
\usepackage{algorithmic}
\usepackage{graphicx}
\usepackage{textcomp}
\usepackage{xcolor}
\usepackage{multicol}
\usepackage{multirow}
\usepackage{booktabs}
\usepackage{bbm}
\usepackage{algorithm}
\usepackage{subcaption}
\author{Jun Xia}
\affiliation{%
  \institution{University of Notre Dame}
  \city{Notre Dame, IN}
  \country{USA}
}
\email{jxia4@nd.edu}

\author{Yiyu Shi}
\affiliation{%
  \institution{University of Notre Dame}
  \city{Notre Dame, IN}
  \country{USA}
}
\email{yshi4@nd.edu}

\AtBeginDocument{%
  \providecommand\BibTeX{{%
    \normalfont B\kern-0.5em{\scshape i\kern-0.25em b}\kern-0.8em\TeX}}}
\begin{document}

\title{Towards Energy-Aware  Federated Learning via MARL: A Dual-Selection Approach for Model and Client }

\renewcommand{\shortauthors}{Trovato et al.}

\begin{abstract}
Although Federated Learning (FL) is promising in knowledge sharing for heterogeneous Artificial Intelligence of
Thing (AIoT) devices, their training performance and energy efficacy are severely restricted in practical battery-driven scenarios due to the ``wooden barrel effect'' caused by the mismatch between homogeneous model paradigms and heterogeneous device capability. 
As a result, due to various kinds of differences among devices, it is hard for existing FL methods to conduct training effectively in energy-constrained scenarios, such as battery constraints of devices.
To tackle the above issues, we propose an energy-aware  FL framework named DR-FL, which considers the energy constraints in both clients and heterogeneous deep learning models to enable energy-efficient FL.
Unlike Vanilla FL, DR-FL adopts our proposed Muti-Agents Reinforcement Learning (MARL)-based  dual-selection  method, which
allows participated devices to  make contributions  to the global model effectively and 
adaptively based on their computing capabilities and energy capacities in a MARL-based manner. 
Experiments conducted with various widely recognized datasets demonstrate that DR-FL has the capability to optimize the exchange of knowledge among diverse models in large-scale AIoT systems while adhering to energy limitations. Additionally, it improves the performance of each individual heterogeneous device's model.

\end{abstract}

\maketitle

\section{Introduction}
The increasing popularity of Artificial Intelligence (AI) techniques, especially  for Deep Learning (DL), accelerates the significant evolution of
Internet of Things (IoT) toward Artificial Intelligence of Things (AIoT), where 
various AIoT devices are equipped with DL models to enable accurate perception and intelligent control \cite{aiot_1}. 
Although   AIoT systems (e.g.,  autonomous driving,  intelligent control \cite{light_d_t}, and healthcare systems \cite{wu_iccad2021,Baghersalimi_bhi2022}) play an important role in various
safety-critical domains, due to both the limited 
classification capabilities of local device models and the restricted access to private local data, 
it is hard to guarantee the training and inference performance of AIoT devices in Federated Learning (FL) \cite{fl_als}, especially when they are powered by batteries and deployed within an uncertain dynamic environment \cite{cui_date2022}. To quickly figure out the training procedure inference perception of devices, more and more large-scale AIoT systems have the aid of cloud computing \cite{zhang_tcad2021}, which has tremendous computing power and flexible device management schemes. 
However, such a cloud-based architecture
still cannot  fundamentally  improve the 
inference accuracy of AIoT devices, 
since they are not allowed to transmit private 
 local data to each other.
Due to  concerns about data privacy, both  
training  and inference performance of local 
models are greatly suppressed.

As a promising  collaborative machine learning paradigm,  FL 
allows local DL model training among various devices without compromising their local data privacy. Instead of sharing local sensitive data among devices,  FL only needs to send gradients or weights of local device models to a cloud server for knowledge aggregation, thus enhancing both the training and inference capability of local models. 
Although FL is promising in  knowledge sharing, 
it faces the problems of both large-scale deployment and quick adaption to dynamic environments, where local models are required to be frequently trained to accommodate an ever-changing world. 
In practice, such problems  are hard to be solved, since vanilla FL methods require that all  devices should have homogeneous local
models with the same architecture.

\begin{figure*}[htbp]
  \centering
\includegraphics[width=\linewidth]{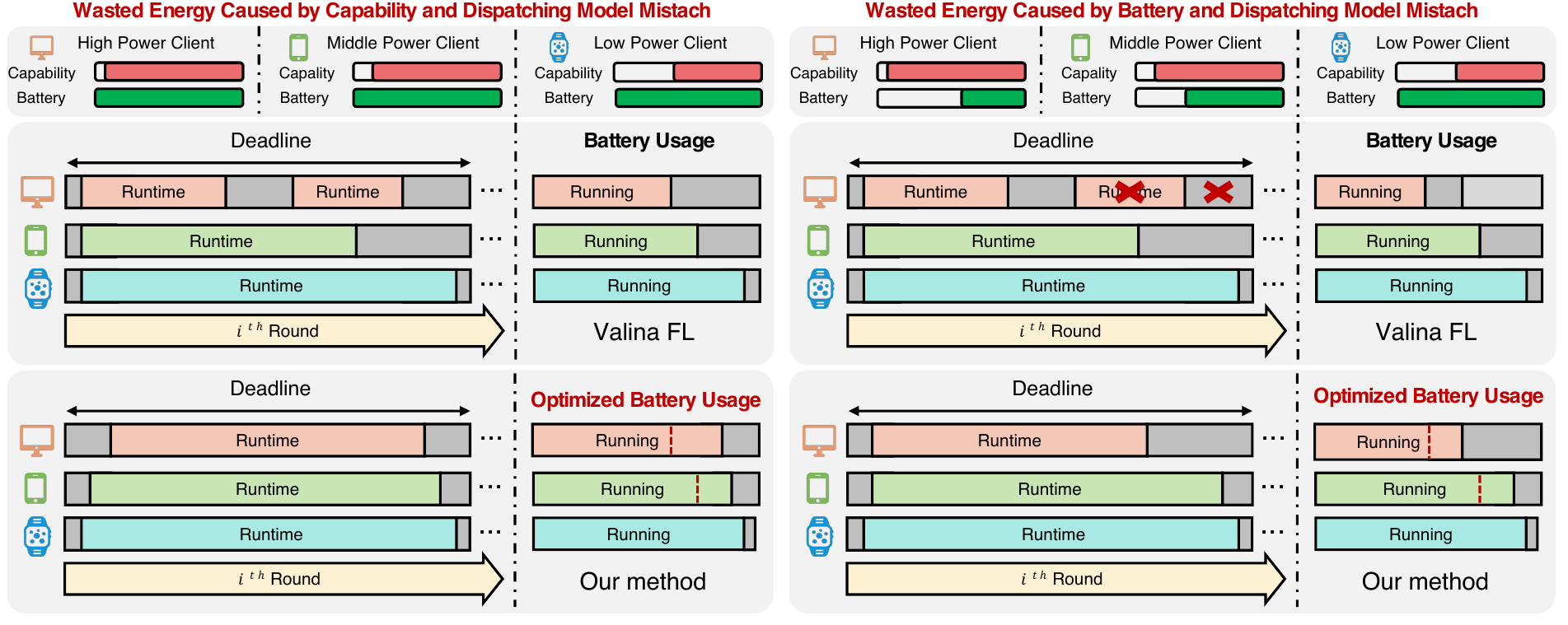}
  \caption{The energy consumption waste of the ``wooden barrel effect'' in Vanilla FL is usually due to the following two reasons, i.e.,  the mismatch between computing power and homogeneous model, and the mismatch between power consumption and homogeneous model. The former uses device energy for waiting time, while the latter uses device energy for useless training time (only enough power to support training but not support communication).}
  \label{motivation}
\end{figure*}

According to the well-known ``wooden barrel effect'' caused by homogeneous assumption as shown in Figure~\ref{motivation}, the energy consumption waste in vanilla FL is usually due to the following two reasons, i.e.,  the mismatch between heterogeneous computing power and homogeneous model, and the mismatch between heterogeneous power consumption and homogeneous model. The former uses device energy for waiting time, while the latter uses device energy for useless training time (only enough power to support training but not support communication). Thus,
such a homogeneous model assumption strongly limits the 
 overall energy efficiency of the entire FL system. This is because energy usages in the entire system are mainly determined by how much power is used in the effective model learning other than waiting or useless training, which consumes energy to wait other than training or communication.

Typically, an  AIoT system  involves 
various types of  
devices with different settings (i.e., computing power and remaining power).
If all  devices have been equipped with homogeneous local models, the inference potential of devices with superior 
computing power will be eclipsed. 
Things become even worse 
when the devices of  AIoT applications  are powered by 
batteries.  
In this case, the devices with less battery energy 
will be reluctant to participate in frequent interactions with the cloud server. Otherwise, if one device runs out of power at an early stage of the FL training, it is hard for the global model to 
achieve an expected 
inference performance.
Meanwhile, 
the overall 
inference performance of the global model will be
strongly deteriorated
due to  the  absence of such an exhausted
 devices
in the following training process. 
Therefore,  \textit{how to fully explore the potential 
of energy-constrained heterogeneous devices to enable high-performance and energy-efficient
FL is becoming a major bottleneck in the  design of an AIoT system.}

Although various heterogeneous FL methods (e.g., 
  HeteroFL \cite{iclr_diao2021}, Scale-FL \cite{scalefl}, PervasiveFL \cite{xia_tcad2022}) and 
 energy-saving techniques \cite{fl_infocom2021,fl_rtss2019}
 have been investigated to address the above issue, most of them focus on either enabling effective knowledge sharing between heterogeneous models or reducing the energy consumption of devices. 
 Based on the coarse-grained 
FedAvg operations, few of the existing FL methods can substantially address the above challenges to quickly adapt to new environments within an energy-constrained scenario. 
Inspired by the concepts of 
BranchyNet \cite{branchynet} and 
multi-agent reinforcement learning  \cite{pami_self_distillation}, in this paper, we propose a novel 
FL  framework named \textit{DR-FL}, which takes both the layer-wise structure information of DL models and the remaining energy of each client into account to enable energy-efficient federated training.
 Unlike the traditional FL method that relies on homogeneous device models, DR-FL maintains a layer-wise global model on the cloud server, while each device only installs a subset layer-wise model according to its computing power and remaining battery. 
In this way, 
all the heterogeneous local models 
 can effectively
 make contributions to the global model  based on their computing capabilities and remaining energy in a MARL-based
manner. 
Meanwhile, by adopting   MARL, DR-FL can not only  make the trade-off between training performance  and energy consumption, 
thus ensuring energy-efficient FL training to accommodate various
energy-constrained environments. 
This paper makes  the following three major contributions: 
\begin{itemize}

\item 

We establish a novel lightweight cloud-based FL framework named DR-FL, which can be easily implemented and enables various heterogeneous DNNs to share knowledge without compromising their data privacy in FL for heterogeneous devices by \textit{layer-wise model aggregation}.
\item  
We propose a \textit{dual-selection approach} based on MARL to control energy-efficient learning from the perspectives of both layer-wise models and participating clients, which can maximize the efficacy of the entire AIoT system. 
\item Experimental results obtained from both simulation and real
test-bed platforms show that, compared with various state-of-the-art approaches, DR-FL can not  only achieve better  inference performance within various  non-IID scenarios, 
but also have superior scalability for large-scale AIoT systems. 
\end{itemize}

The rest of this paper is organized as follows.
Section \ref{rel} discusses related work on heterogeneous FL  and energy-aware FL training. 
After giving the preliminaries of FL and multi-agent reinforcement learning in section \ref{prel}, section \ref{app} 
details our proposed DR-FL method. 
Section \ref{exp} presents experimental results on well-known benchmarks. Finally, section \ref{con} concludes the paper.

\section{Related Work}
\label{rel}

Although FL is good at knowledge sharing without compromising the data privacy of devices in AIoT system design, due to the homogeneous assumption that all the involved devices should have local
DL models with the same architecture,  vanilla FL methods inevitably suffer from the problems of 
low inference accuracy and invalid  energy consumption, thus impeding the deployment of FL methods in 
large-scale AIoT system designs \cite{fl_cst2021,fl_cst2021b,xia_tcad2022,zhu_icml2021}, especially for  non-IID scenarios with constrained energy limitation.

To facilitate collaborative learning among different device models, many works have thoroughly examined numerous solutions. These solutions may be broadly categorized into two types: subnetwork aggregation-based methods and knowledge distillation-based methods.
Subnetwork aggregation-based methods aim to facilitate knowledge aggregation by aggregating subnetworks of local device models. This approach enables the sharing of knowledge among diverse device models.
 For instance, Diao et al. \cite{iclr_diao2021} presented
 an effective heterogeneous FL framework named HeteroFL, 
 which can train heterogeneous local models with varying computation complexities but
 still produce a single global inference model, assuming that 
device models are subnetworks of the global model. 
By integrating FL and width-adjustable slimmable neural networks, Yun et al. \cite{ton_2023} proposed a novel learning framework named ScaleFL, 
which jointly utilizes superposition coding for global model aggregation and training for updating local models.
 In \cite{xia_tcad2022}, Xia et al. developed a novel   framework named PervasiveFL, which utilizes a small 
 uniform model (i.e., ``modellet'') to enable heterogeneous FL in AIoT systems. 
 Although all the above heterogeneous FL methods are promising, most of them focus on improving inference performance of local models. Few of them take the issues of real-time training and energy efficiency into account.

Since a large-scale FL-based AIoT application typically involves a variety of devices that are powered by batteries, how to conduct energy-efficient FL training is becoming an important issue \cite{fl_twc2022,aiot}. To address this issue, various methods have been investigated to reduce the energy consumed by FL training and device-server communication.  
For example, 
Hamdi et al. \cite{fl_iotj2022} studied the FL deployment problem 
in an energy-harvesting wireless network, where a certain number of users may be unable to participate in FL due to interference and energy constraints.  They  formalized such a deployment scenario as a 
joint energy management and user scheduling problems over wireless systems,
and solved it efficiently. 
 In \cite{fl_icc2020},  Sun et al. presented
  an online energy-aware dynamic worker scheduling policy, 
  which can maximize the average number of workers scheduled for gradient update 
  under a long-term energy constraint. 
In \cite{fl_twc2021}, Yang et al. formulated the energy-efficient transmission and computation resource allocation for FL over wireless communication networks as 
a joint learning and communication problem. To minimize system energy consumption under a latency constraint, they presented an iterative algorithm that can derive 
optimal solutions considering various factors (e.g.,  bandwidth allocation, power control, computation frequency, and learning accuracy).
Although all the above energy-saving methods can effectively reduce energy consumption in both FL training and communication, few of them can guarantee the training time requirement of FL training within a complex dynamic environment.

To the best of our knowledge, DR-FL is the first attempt to investigate the dual selection by both layer-wise models and the participated clients based on MARL to enable fine-grained heterogeneous FL, where heterogeneous devices can adaptively and efficiently make contributions to the global model based on their computing capabilities and remaining energy.
DR-FL surpasses other advanced heterogeneous FL approaches by optimizing information transfer among different models with limited energy resources and enhancing both the performance of individual devices and the energy efficiency of the entire FL system.
\section{Preliminaries\label{prel}} 
\subsection{Federated Learning}
With the prosperity of distributed machine learning technologies \cite{dml_csur}, privacy-aware FL is proposed to effectively solve the problem of data silos, where multiple AIoT devices can achieve knowledge sharing without leaking their data privacy.
Since the physical environment is volatile (i.e., high latency network and unstable connection) in real AIoT scenarios, Vanilla FL randomly selects a number of AIoT devices for each communication round of training a homogeneous DNN model.
Suppose there are $N$ devices selected at the  $t^{th}$ communication round in FL. After the $t^{th}$ communication round, the update process of each device model is defined  as follows
 \begin{equation}
    \mathbb{W}_{t+1}^{n} \gets \mathbb{W}_{t}^{n} - \eta \nabla \mathbb{W}_{t}^{n} ,
\label{weight update}
 \end{equation}
 where $\mathbb{W}_{t}^{n}$ and $\mathbb{W}_{t+1}^{n}$ represent the global models at round $t$ and round $t+1$ in the $n^{th}$ device, respectively. $\eta$ indicates the learning rate and $\nabla \mathbb{W}_{t}^{n}$ is the gradient obtained by the $n^{th}$ device model after the $t^{th}$ training round. 
 To preserve the data privacy of local devices, at the end of each communication round, FL uploads each device's weight differences (i.e., model gradients) instead of updated NN models to the cloud for aggregation.
 After gathering the gradients from all the participating devices, the cloud updates the parameters of the shared-global model based on the Fedavg \cite{FL-google} algorithm, which is defined as follows:
 \begin{equation}
    \mathbb{W}_{t+1}\gets \mathbb{W}_{t}+ \frac{\sum_{n=1}^N\mathbb{L}_{n}\nabla \mathbb{W}_{t}^{n}}{N},
\label{weight aggregration}
 \end{equation}
 where $\frac{\sum_{n=1}^N\nabla \mathbb{W}_{t}^{n}}{N}$ denotes the average gradient of $N$ participating devices in  communication round $t$, $\mathbb{W}_{t}$
and $\mathbb{W}_{t+1}$ represent the  global models after $t^{th}$ and $t+1^{th}$ communication round, respectively, and $\mathbb{L}_n$ means the training data size of device $n$.
 Although vanilla FL methods (e.g., FedAvg) perform remarkably in distributed machine learning, they cannot be directly applied to AIoT scenarios. This is because the heterogeneous AIoT devices will lead to different training speeds for vanilla FL, resulting in additional energy waste, which is unacceptable for an efficient energy-constrained system.

 \subsection{Multi-Agent Reinforcement Learning }

 Cooperative Multi-Agent Reinforcement Learning (MARL) involves training a group of \(N\) agents to generate optimal actions that result in the highest possible team rewards. At each timestamp \(t\), each agent \(n\) (where \(1 \leq n \leq N\)) observes its state \(s_t^n\) and chooses an action \(a_t^n\) based on \(s_t^n\). 
 Once all agents have finished their actions, the team is given a collective reward \(r_t\) and moves on to the next state \(s_{t+1}^n\). The objective is to optimize the overall predicted discounted reward \(R = \sum_{t=1}^T \gamma r_t\) by choosing the best behaviours for each agent. Here, \(\gamma \in [0, 1]\) represents the discount factor for reward.

Recently, QMIX \cite{qmix} has emerged as a promising solution for jointly training agents in cooperative MARL. 
In QMIX, each agent \(n\) employs a Deep Neural Network (DNN) to infer its actions. This DNN implements the \(Q\)-function \(Q^\theta(s, a) = E[R_t|s_t^n = s, a_t^n = a]\), where \(\theta\) represents the parameters of the DNN, and \(R_t = \sum_{i=t}^T \gamma r_i\) is the total discounted team reward received at \(t\). During MARL execution, each agent \(n\) selects the action \(a^*\) with the highest \(Q\)-value (i.e., \(a^* = \arg\max_a Q^\theta(s_n, a)\)).
To train the QMIX, a replay buffer is employed to store transition tuples \((s_t^n, a_t^n, s_{t+1}^n, r_t)\) for each agent \(n\). The joint \(Q\)-function, \(Q_{\text{tot}}(\cdot)\), is represented as the element-wise summation of all individual \(Q\)-functions (i.e., \(Q_{\text{tot}}(s_t, a_t) = \sum_n Q^\theta_n(s_t^n, a_t^n)\)), where \(s_t = \{s_t^n\}\) and \(a_t = \{a_t^n\}\) are the states and actions collected from all agents \(n \in N\) at timestamp \(t\). The agent DNNs can be recursively trained by minimizing the loss \(L = E_{s_t,a_t,r_t,s_{t+1}}[y_t - Q_{\text{tot}}(s_t, a_t)]^2\), where \(y_t = r_t + \gamma \sum_n \max_a Q^{\theta'}_n(s_{t+1}^n, a)\) and \(\theta'\) represents the parameters of the target network, which are periodically copied from \(\theta\) during the training phase.


\section{Method}
\label{app}
\subsection{Problem Formulation}
Given an energy-limited federated learning (FL) system consisting of a cloud server and $N$ diverse AIoT devices, denoted as $D = \left \{ D_{1}, ..., D_{n}, ..., D_{N} \right \}$. These diverse AIoT devices can be categorized into three separate categories based on their computing capability: small, medium, and large. The terms small, medium, and large refer to the level of computational and storage capabilities available on the devices.
Three crucial aspects, namely running time, energy consumption, and model correctness, substantially impact the overall performance of the FL system discussed in this paper. The running time of the FL system directly impacts the training efficiency in a real-world scenario. 
Furthermore, energy consumption is a crucial aspect, especially for AIoT devices that rely on limited energy resources. 
Finally, the accuracy of the model guarantees that the system generates reliable and valuable predictions. 
Hence, to enhance the FL system's overall efficiency, achieving a harmonious equilibrium among three key components is imperative.




\textbf{Running Time Model:}
Considering the differences in network delay and computing resources of heterogeneous AIoT devices, the energy-constrained FL system aims to minimize the total running time $T_{all}$ among all the devices, which is shown as
\vspace{-0.05in}
\begin{equation}
T_{all}=\max_{\forall n} T_{all}^{D_{n}}.
\vspace{-0.05in}
\end{equation}
Let $T_{com}^{D_{n}}$ and $T_{tra}^{D_{n}}$ be the communication time of the device $D_{n}$ and the training time of the layer-wise model on device $D_{n}$, respectively. Note that due to the abundant computing resources in the cloud server, its running time is negligible compared to that on devices. The total running time for each device $T_{all}^{D_{n}}$ is defined as
\vspace{-0.04in}
\begin{equation}
T_{all}^{D_{n}}= T_{com}^{D_{n}} + T_{tra}^{D_{n}}.
\vspace{-0.04in}
\end{equation}
Here, the communication time for each device $T_{com}^{D_{n}}$ can be regarded as the ratio of the size of a model $S_{D_{n}}$ with different layers and the speed of bandwidth  $V_{net}$. 
Since the training time of each device $T_{tra}^{D_{n}}$ is determined by the computation capability of local devices $C_{D_{n}}$, the training data size in a device $L_{D_{n}}$, we formalize communication time $T_{com}^{D_{n}}$ and training time $T_{tra}^{D_{n}}$ as
\vspace{-0.025in}
\begin{equation}
\begin{matrix}
T_{com}^{D_{n}} = \frac{S_{D_n}}{V_{net}}, \qquad 
T_{tra}^{D_{n}} = \frac{L_{D_{n}}}{C_{D_{n}}},
\end{matrix}
\label{time_all_calc}
\vspace{-0.025in}
\end{equation}
where $O_{D_{n}}$ is reflected by the computation capability of the device $C_{D_{n}}$. Assuming that the network transmission speed can be kept relatively stable.


\textbf{Energy Consumption Model:}
The energy consumed by the overall FL system plays an important role in ensuring the system operates smoothly. The calculation of the total remaining energy  can be expressed as
\vspace{-0.04in}
\begin{equation}
E_{all} = \sum_{n=1}^{N} \left ( E_{remain}^{D_{n}} - E_{tra}^{D_{n}}-E_{com}^{D_n} \right ).
\label{eq:calc_energy}
\vspace{-0.04in}
\end{equation}
Note that both training and communication energy consumption are all decided by two factors, i.e., the size of the training model and the power mode of AIoT devices.
The training energy consumption $E_{tra}^{D_{n}}$ and communication energy consumption $E_{com}^{D_{n}}$ of device $D_n$ are calculated as
\vspace{-0.03in}
\begin{equation}
\begin{matrix}
E_{tra}^{D_{n}}= P_{train} \times T_{tra}^{D_{n}}, \qquad 
E_{com}^{D_{n}}= P_{com} \times T_{com}^{D_{n}},
\end{matrix}
\vspace{-0.03in}
\label{energy_calc}
\end{equation}
where $P_{train}$ is the energy consumption per unit training time, and $P_{com}$ is the energy consumption per unit network transmission time. 
Note that since actual energy consumption is intrinsically related to the size of the trained model, variations in the size of the model lead to fluctuations in the energy consumed during both the training and communication processes.
Therefore, it is of utmost importance to consider these energy dynamics when addressing the optimization model.

\textbf{Model Accuracy:}
The appropriate utilization of heterogeneity in heterogeneous models and devices to improve the performance of aggregated models is an urgent issue that must be addressed in the field of Federated Learning (FL).
In addition, the use of energy-limited federated learning is hindered by the presence of resource-constrained heterogeneous AIoT devices that are involved in data aggregation.
Based on the findings of the study referenced as \cite{fl_rtss2019}, it can be inferred that the accuracy of heterogeneous models is directly related to the number of successful aggregations for each device. In other words, the more aggregated models participating in each round, the higher the accuracy of the model inference.
Nevertheless, the issue in designing a Federated Learning (FL) framework lies in the selection of aggregation devices in an energy-constrained environment to enhance model accuracy, considering that devices waste energy during each cycle of aggregation.

\textbf{Optimization Objective:}
Taking energy information into account, an optimization model is proposed for energy-constrained FL. 
This model seeks to achieve a compromise between three objectives: minimizing the overall running time $T_{all}$, maximizing the model accuracy $M_{acc}$ while adhering to a constraint on total energy consumption $E_{all}$. The constraint is defined as follows: 
\begin{equation}
\begin{matrix}
\min T_{all}, \ \quad
\max  M_{acc}, \\
\text{s.t.} \ E_{all} \leq E.
\end{matrix}
\label{eq:optimization_model_compact}
\end{equation}
Here, $E$ represents the energy allocation of a FL system.
\begin{figure*}[htbp]
  \centering
  \includegraphics[width=0.85\linewidth]{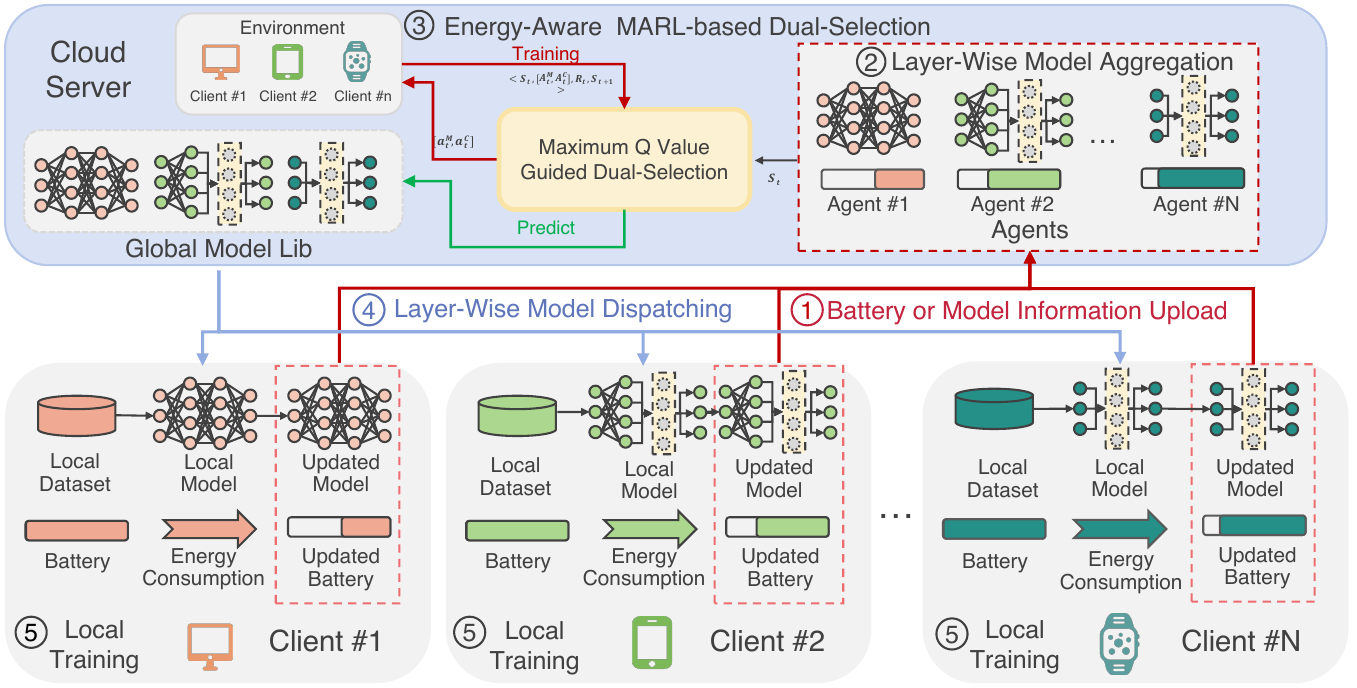}
  \vspace{-0.05in}
  \caption{Framework and workflow of our method.}
  \label{framework}
  \vspace{-0.15in}
\end{figure*}
\subsection{Workflow   of DR-FL}

DR-FL involves the collaboration of diverse AIoT devices and a cloud server to optimize the performance of different layer-wise models implemented on edge devices. Prior to training, all devices involved in DR-FL will initialize and install a layer-wise model, which is a subset layer of the global model stored on the cloud server. 
Subsequently, the cloud server transmits a segment of the comprehensive model to AIoT devices for localized training. After completing the training process on the local device, DR-FL carries out layer-wise model aggregation on the cloud server. It should be noted that hot-plug AIoT devices are allowed with DR-FL. These newly connected devices only receive the parameters of the global model from the cloud server. Figure~\ref{framework} depicts the process of the DR-FL, which comprises five serial steps.

{\bf  Step 1 (Battery or Model Information Upload):} 
During the initialization step of DR-FL, each device intending to participate in FL should upload its device information to the cloud, which includes the power, computing, and storage capabilities of devices and the overclocking potential of models.
In subsequent steps, this information is used for energy-aware dual-selection for the layer-wise model and client to optimize the entire system's energy efficiency.

{\bf Step 2 (Layer-Wise Model Aggregation):} 
After receiving the participating devices' local model gradients, this step will layer-align averaging (\textit{The same parts of the network will be aggregated.}) such gradients and use the previous round global model stored on the server to construct a new global model. 

{\bf Step 3 (Energy-Aware MARL-based Dual-Selection):} 
Then, to prevent selected devices from dropping out of the FL process due to energy limitations, we design a MARL-based selector that can choose an appropriate model for each AIoT device based on its remaining energy and computing capabilities, which can not only improve the efficiency of the device resource usage but also ensure their active participation in FL (see more details in Section~\ref{sec:dual_sel}).
Furthermore, apart from selecting a layer-wised model for each AIoT device, the selector can also adjust the computing capability of AIoT devices, aiming to achieve a trade-off between energy consumption and computing efficiency.

{\bf  Step 4 (Layer-Wise Model Dispatching):} 
Based on an energy-aware MARL-based dual-selection strategy, the cloud server dispatches part of the global model parameters to each heterogeneous AIoT device.

{\bf Step 5 (Local Training):} 
Based on the received global model parameters, each heterogeneous AIoT device builds an initial local model (i.e., layer-wise model), which is trained using cross-entropy loss based on local training samples to obtain the gradients of the local model for gradient upload.

DR-FL repeats all five steps above until the global model and all its local models converge.

\subsection{ Dual-Selection for Local Model and Client}\label{sec:dual_sel}
\begin{figure}[htbp]
  \centering
  \includegraphics[width=\linewidth]{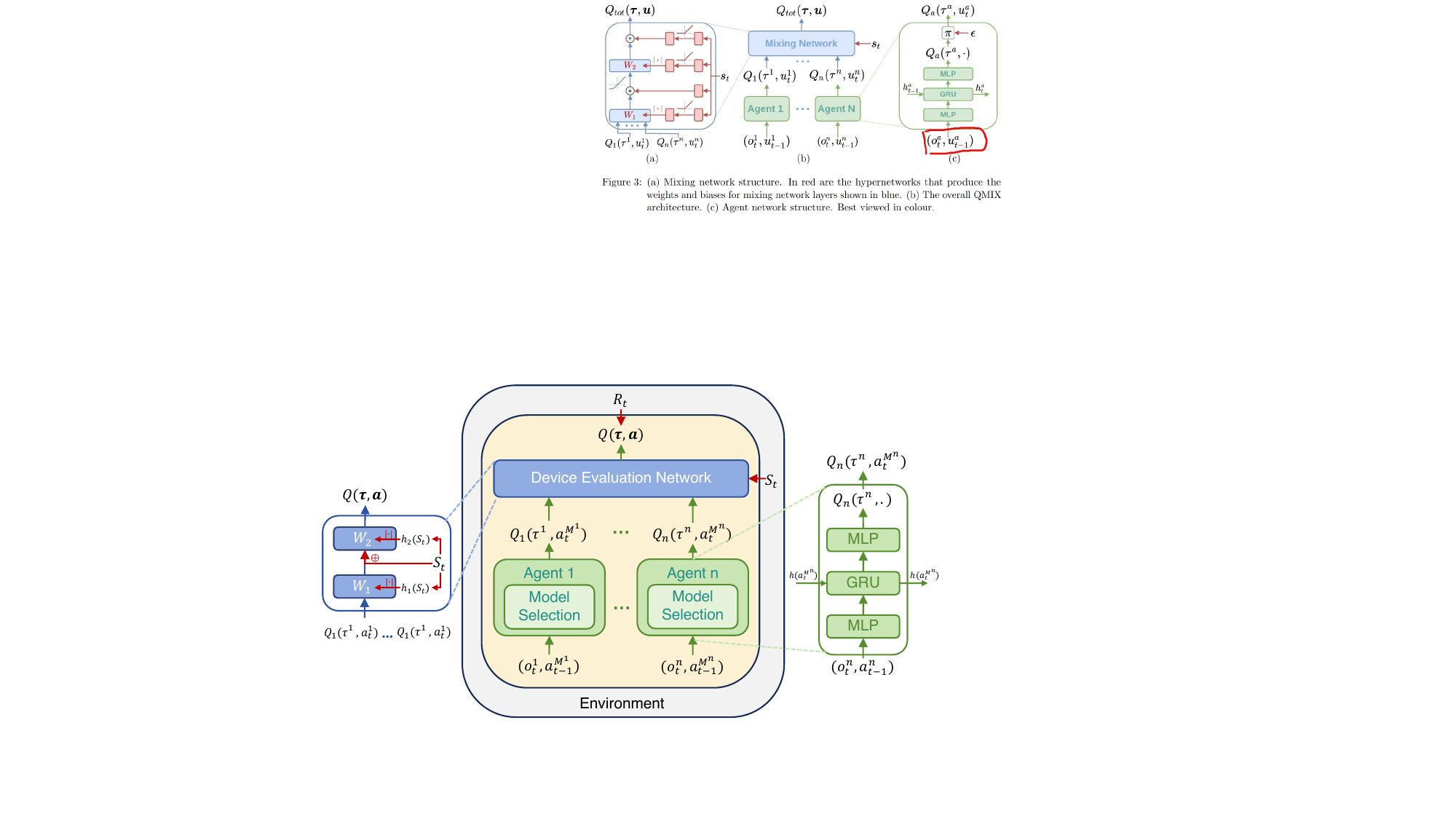}
  \caption{Maximum Q Value Guided Dual-selection. There are two networks here, i.e., the model selection network and the device evaluation network. 
The model selection network is calculated through the value $O$ observed by the agent from the environment and the action set $A_{t-1}$ of the previous round, thereby obtaining the latest action and its corresponding Q value. 
The device evaluation network obtains the Q values of all devices and then uses the hybrid network to combine all Q values and the current timestamp state $S_{t}$ through a two-layer weight matrix into an overall Q value $Q_{tot}$.  Then, the network uses the discounted rewards given by the environment for MARL, thereby multi-agents can obtain their own rewards from the environment. $h$ means the MLP for extracting deep representations of states or actions. $|\cdot|$ means the dot product.}
  \vspace{-0.2in}
  \label{qmixing}
\end{figure}
\subsubsection{MARL Training Process:}
In our DR-FL, each device uses an energy-aware MARL-based dual-selection method to select the participated device and the layers of its corresponding local model running on devices. 
To better capture connections between long-term/short-term rewards and strategies, each MARL is designed with two Multi-Layer Perceptions (MLP) and a Gated Recurrent Unit (GRU) \cite{GRU}, respectively, as shown in Figure~\ref{qmixing}.
During the training procedure of MARL, each agent acquires its current state $S_{t}$ and selects an action $a_t^n$ for each client.
 Based on both client selection and layer-wise model considerations, the central server computes team rewards by considering the validation accuracy improvement of the global model $M_{acc}$, the total runtime $T_{all}$, the computation capabilities $C$ and the remaining energy of each device $E_{all}$.
 The MARL agents are then trained with the QMIX algorithm \cite{qmix} to maximize the system rewards (See the design details in Section~\ref{sec:reward_calc}).

\subsubsection{MARL Agent State Design: }
The state of each MARL agent $D_{n}$ is comprised of three components: the remaining energy $E_{all}^{D_{n}}$, the computation capability of each communication round $C_{D_{n}}$, and the size of the local training dataset $L_{D_{n}}$.
At each training round $t$, each agent initially conducts the training procedure and transmits its gradients to the central server. 
Furthermore, to estimate the current training and communication delays at client device $n$, each MARL agent is equipped with a record of training latency $T_{tra}^{D_n}$ and communication latency $T_{com}^{D_n}$, where $T_{tra}^{D_n}$ and $T_{com}^{D_n}$ denote the latency in local training and model uploading for agent $n$ during the communication round $t$. 
As shown in Figure~\ref{qmixing}, the parameter $\tau$ represents the trajectory of historical data from training, and $h$ represents the MLP layer for knowledge extraction.
Moreover, each MARL agent $n$ also calculates the energy consumption of training and communication based on Equation~\ref{energy_calc}. 
This inclusion is crucial as the energy costs contribute to the overall energy cost, while the remaining energy of the agent influences both training latency and model accuracy. 
The state vector \(s_t^n\) of agent \(n\) in communication round \(t\) is defined as:
\begin{equation}
s_t^n = [L_t^n, C_{D_n}, E_{D_n}, t].
\label{state}
\end{equation}

Finally, to decrease storage overhead and accelerate the speed of agent convergence, all MLPs and GRUs within the MARL agents share their weights.

\subsubsection{Agent Action Design:}
Given the input state shown in Equation~\ref{state}, each MARL agent $n$ determines which layers of the local model should be used for the local training process on each device.
 Specifically, the MARL agent will generate  $Q$ values for the current action set $[a^0, \ldots, a^M]$, where $M$ represents the number of model selections available to the client.
 Note that when the selected action is zero, the client device will run the first model, and when the selected action is $M$, the client will not participate in the FL. 
 After selecting the layer-wise model for each heterogeneous device, all the Q values obtained by the agents will select the device with the highest Q value through the Top-K algorithm to participate in the FL process.

\subsubsection{Reward Function Design:}\label{sec:reward_calc}
To optimize the objective described in Equation~\ref{eq:optimization_model_compact}, the reward function should reflect the changes in the model accuracy, processing latency (training,  communication and waiting latency),  and processing energy consumption after executing the dual-selection strategy generated by  MARL agents. 
The reward \(r_t\) at training round \(t\) is defined as follows:

\begin{equation}
r_t = w_1\cdot(M_{Acc}^t - M_{Acc}^{t-1} )- w_2 \cdot (E_{all}^{t-1} - E_{all}^{t}) - w_3 \cdot \max_{1 \leq n \leq N} T_{all}^{t, n}.
\label{eq:reward_calc}
\end{equation}

Here, $\max_{1 \leq n \leq N} T_{all}^{t, n}$ represents the total time needed for local training of all selected devices. The MARL agents utilize the evaluation accuracy calculated by a small tiny dataset on the cloud server to select the layer-wise model that will be dispatched to the local device and continue the local training and upload their model updates. Moreover, $w_1$, $w_2$, and $w_3$ \footnote{We used $w1 = 1000, w2= 0.01, w3=1$ in our experiments.} are the norm ratios to control all the reward plays the same role in the entire reward.
$E_{all}^t$ is the total remaining energy  of $t^{th} $ communication round as defined in Equation~\ref{eq:calc_energy}. The MARL agents are trained using QMIX as described in Figure~\ref{qmixing}.

\begin{table*}[htbp]
\centering
  \caption{Test accuracy (\%) comparison for different models and dataset settings under specific energy constraints with 40 clients.}
  \label{test_acc_energy}
\footnotesize
\begin{tabular}{|c|ccccccccc|}
\hline
Dataset      & \multicolumn{9}{c|}{CIFAR10}                                                                                        \\ \hline
Methods      & \multicolumn{3}{c|}{HeteroFL \cite{iclr_diao2021}}                                                                                              & \multicolumn{3}{c|}{ScaleFL \cite{scalefl}}                                                                                   & \multicolumn{3}{c|}{DR-FL (Ours)}                                                                                                 \\ \hline
Distribution & \multicolumn{1}{c|}{$\alpha$=0.1} & \multicolumn{1}{c|}{$\alpha$=0.5}          & \multicolumn{1}{c|}{$\alpha$=1.0}    & \multicolumn{1}{c|}{$\alpha$=0.1} & \multicolumn{1}{c|}{$\alpha$=0.5} & \multicolumn{1}{c|}{$\alpha$=1.0}  & \multicolumn{1}{c|}{$\alpha$=0.1}          & \multicolumn{1}{c|}{$\alpha$=0.5}          & \multicolumn{1}{c|}{$\alpha$=1.0 }             \\ \hline
Model\_1      & \multicolumn{1}{c|}{$30.46\pm 1.10$} & \multicolumn{1}{c|}{46.11 $\pm$ 3.32}          & \multicolumn{1}{c|}{65.23 $\pm$ 1.45}  & \multicolumn{1}{c|}{29.25 $\pm$ 1.17} & \multicolumn{1}{c|}{54.44 $\pm$ 0.87} & \multicolumn{1}{c|}{58.15 $\pm$ 4.32}   & \multicolumn{1}{c|}{\textbf{58.69 $\pm$ 0.73}} & \multicolumn{1}{c|}{\textbf{59.01 $\pm$ 0.85}} & \multicolumn{1}{c|}{\textbf{76.46 $\pm$ 0.12} }
\\ \hline
Model\_2      & \multicolumn{1}{c|}{48.41 $\pm$ 1.24} & \multicolumn{1}{c|}{62.55 $\pm$ 3.45}          & \multicolumn{1}{c|}{62.10 $\pm$ 3.24} & \multicolumn{1}{c|}{41.66 $\pm$ 5.43} & \multicolumn{1}{c|}{55.46 $\pm$ 3.87} & \multicolumn{1}{c|}{71.48 $\pm$ 1.23}    & \multicolumn{1}{c|}{\textbf{65.31 $\pm$ 1.54}} & \multicolumn{1}{c|}{\textbf{75.93 $\pm$ 0.62}} & \multicolumn{1}{c|}{\textbf{77.43 $\pm$ 2.77}} 
\\ \hline
Model\_3      & \multicolumn{1}{c|}{34.85 $\pm$ 5.79} & \multicolumn{1}{c|}{65.01 $\pm$ 1.79 } &  \multicolumn{1}{c|}{\textbf{74.78 $\pm$ 2.76}}  & \multicolumn{1}{c|}{39.92 $\pm$ 2.75} & \multicolumn{1}{c|}{60.07 $\pm$ 0.68} & \multicolumn{1}{c|}{70.83 $\pm$ 1.43}    & \multicolumn{1}{c|}{\textbf{72.71 $\pm$ 0.58}} & \multicolumn{1}{c|}{\textbf{70.64 $\pm$ 1.40}}          & \multicolumn{1}{c|}{71.54 $\pm$ 1.54} \\ \hline
Model\_4      & \multicolumn{1}{c|}{45.26 $\pm$ 3.68} & \multicolumn{1}{c|}{69.65 $\pm$ 2.99} & \multicolumn{1}{c|}{\textbf{75.14 $\pm$ 1.13}}  & \multicolumn{1}{c|}{46.59 $\pm$ 3.43} & \multicolumn{1}{c|}{\textbf{70.60 $\pm$ 4.54}} & \multicolumn{1}{c|}{73.90 $\pm$ 1.17}     & \multicolumn{1}{c|}{\textbf{70.76 $\pm$ 1.30}} & \multicolumn{1}{c|}{69.37 $\pm$ 0.45}          & \multicolumn{1}{c|}{72.27 $\pm$ 1.73} 
\\ \hline\hline
Dataset      & \multicolumn{9}{c|}{CIFAR100}                                                                                                                                                                                                                                                                                                                                                  \\ \hline
Methods      & \multicolumn{3}{c|}{HeteroFL \cite{iclr_diao2021}}                                                                                              & \multicolumn{3}{c|}{ScaleFL \cite{scalefl}}                                                                                   & \multicolumn{3}{c|}{DR-FL (Ours)}                                                                                                 \\ \hline
Distribution & \multicolumn{1}{c|}{$\alpha$=0.1} & \multicolumn{1}{c|}{$\alpha$=0.5}          & \multicolumn{1}{c|}{$\alpha$=1.0}   & \multicolumn{1}{c|}{$\alpha$=0.1} & \multicolumn{1}{c|}{$\alpha$=0.5} & \multicolumn{1}{c|}{$\alpha$=1.0}  & \multicolumn{1}{c|}{$\alpha$=0.1}          & \multicolumn{1}{c|}{$\alpha$=0.5}          & \multicolumn{1}{c|}{$\alpha$=1.0}                   \\ \hline
Model\_1      & \multicolumn{1}{c|}{11.86 $\pm$ 0.78} & \multicolumn{1}{c|}{22.56 $\pm$ 2.13}          & \multicolumn{1}{c|}{25.66 $\pm$ 1.13}  & \multicolumn{1}{c|}{13.14 $\pm$ 1.96} & \multicolumn{1}{c|}{21.39 $\pm$ 1.59} & \multicolumn{1}{c|}{17.58 $\pm$ 0.43}     & \multicolumn{1}{c|}{\textbf{26.25 $\pm$ 0.23}} & \multicolumn{1}{c|}{\textbf{33.59 $\pm$ 3.32}} & \multicolumn{1}{c|}{\textbf{39.65 $\pm$ 1.35}} \\ \hline
Model\_2      & \multicolumn{1}{c|}{16.33 $\pm$ 3.34} & \multicolumn{1}{c|}{25.98 $\pm$ 1.72}          & \multicolumn{1}{c|}{28.68 $\pm$ 0.57}  & \multicolumn{1}{c|}{12.67 $\pm$ 2.13} & \multicolumn{1}{c|}{28.77 $\pm$ 4.33} & \multicolumn{1}{c|}{29.84 $\pm$ 1.39}    & \multicolumn{1}{c|}{\textbf{17.83 $\pm${0.75}}} & \multicolumn{1}{c|}{\textbf{39.50 $\pm$ 1.08}} & \multicolumn{1}{c|}{\textbf{33.55 $\pm$ 0.45}} \\ \hline
Model\_3      & \multicolumn{1}{c|}{14.18 $\pm$ 0.29} & \multicolumn{1}{c|}{31.99 $\pm$ 0.53}          & \multicolumn{1}{c|}{31.31 $\pm$ 3.34} & \multicolumn{1}{c|}{17.12 $\pm$ 2.88} & \multicolumn{1}{c|}{30.04 $\pm$ 1.91} & \multicolumn{1}{c|}{\textbf{33.92 $\pm$ 2.34}}   & \multicolumn{1}{c|}{\textbf{26.46 $\pm$ 0.24}} & \multicolumn{1}{c|}{\textbf{32.10 $\pm$ 1.12}} & \multicolumn{1}{c|}{33.40 $\pm$ 0.13} \\ \hline
Model\_4      & \multicolumn{1}{c|}{15.66 $\pm$ 0.78} & \multicolumn{1}{c|}{29.33 $\pm$ 0.85}          & \multicolumn{1}{c|}{35.44 $\pm$ 1.54}  & \multicolumn{1}{c|}{19.24 $\pm$ 1.22} & \multicolumn{1}{c|}{30.29 $\pm$ 1.03} & \multicolumn{1}{c|}{33.23 $\pm$ 1.32}    & \multicolumn{1}{c|}{\textbf{22.55 $\pm$ 0.73}} & \multicolumn{1}{c|}{\textbf{32.55 $\pm$ 1.45}} & \multicolumn{1}{c|}{\textbf{33.80 $\pm$ 1.25}} \\\hline\hline
Dataset      & \multicolumn{9}{c|}{SVHN}   
   \\ \hline
Methods      & \multicolumn{3}{c|}{HeteroFL \cite{iclr_diao2021}}                                                                                              & \multicolumn{3}{c|}{ScaleFL \cite{scalefl}}                                                                                   & \multicolumn{3}{c|}{DR-FL (Ours)}                                                                                                 \\ \hline
Distribution & \multicolumn{1}{c|}{$\alpha$=0.1} & \multicolumn{1}{c|}{$\alpha$=0.5}          & \multicolumn{1}{c|}{$\alpha$=1.0}    & \multicolumn{1}{c|}{$\alpha$=0.1} & \multicolumn{1}{c|}{$\alpha$=0.5} & \multicolumn{1}{c|}{$\alpha$=1.0}  & \multicolumn{1}{c|}{$\alpha$=0.1}          & \multicolumn{1}{c|}{$\alpha$=0.5}          & \multicolumn{1}{c|}{$\alpha$=1.0}                 \\ \hline
Model\_1      & \multicolumn{1}{c|}{60.08 $\pm$ 3.23} & \multicolumn{1}{c|}{46.02 $\pm$ 3.32}          & \multicolumn{1}{c|}{60.38 $\pm$ 1.39} & \multicolumn{1}{c|}{47.90 $\pm$ 0.53} & \multicolumn{1}{c|}{85.79 $\pm$ 2.22} & \multicolumn{1}{c|}{\textbf{88.91 $\pm$ 1.11}}     & \multicolumn{1}{c|}{\textbf{67.19 $\pm$ 0.32}} & \multicolumn{1}{c|}{\textbf{91.58 $\pm$ 0.21}} & \multicolumn{1}{c|}{68.78 $\pm$ 1.33}  \\ \hline
Model\_2      & \multicolumn{1}{c|}{65.11 $\pm$ 4.32} & \multicolumn{1}{c|}{54.83 $\pm$ 1.28}          & \multicolumn{1}{c|}{68.90 $\pm$ 2.87}  & \multicolumn{1}{c|}{50.26 $\pm$ 2.21} & \multicolumn{1}{c|}{86.82 $\pm$ 2.51} & \multicolumn{1}{c|}{85.16 $\pm$ 4.13}    & \multicolumn{1}{c|}{\textbf{79.86 $\pm$ 0.87}} & \multicolumn{1}{c|}{\textbf{85.30 $\pm$ 1.19}} & \multicolumn{1}{c|}{\textbf{91.72  $\pm$ 0.94}}  \\ \hline
Model\_3      & \multicolumn{1}{c|}{65.93 $\pm$ 4.56} & \multicolumn{1}{c|}{69.20 $\pm$ 4.19}          & \multicolumn{1}{c|}{75.97 $\pm$ 1.84}  & \multicolumn{1}{c|}{76.73 $\pm$ 2.23} & \multicolumn{1}{c|}{84.91 $\pm$ 0.68} & \multicolumn{1}{c|}{88.70 $\pm$ 3.25}     & \multicolumn{1}{c|}{\textbf{91.47 $\pm$ 0.17}} & \multicolumn{1}{c|}{\textbf{88.61 $\pm$ 1.72}} & \multicolumn{1}{c|}{\textbf{93.45 $\pm$ 0.37}}  \\ \hline
Model\_4      & \multicolumn{1}{c|}{66.31 $\pm$ 3.09} & \multicolumn{1}{c|}{71.34 $\pm$ 0.79}          & \multicolumn{1}{c|}{76.14 $\pm$ 1.90}  & \multicolumn{1}{c|}{55.27 $\pm$ 3.23 } & \multicolumn{1}{c|}{86.10 $\pm$ 3.56} & \multicolumn{1}{c|}{92.47 $\pm$ 0.51}    & \multicolumn{1}{c|}{\textbf{91.11 $\pm$ 1.32}} & \multicolumn{1}{c|}{\textbf{89.26 $\pm$ 0.75}} & \multicolumn{1}{c|}{\textbf{92.78 $\pm$ 0.54}}  \\\hline\hline
Dataset      & \multicolumn{9}{c|}{Fashion-MNIST}   
   \\ \hline
Methods      & \multicolumn{3}{c|}{HeteroFL \cite{iclr_diao2021}}                                                                                              & \multicolumn{3}{c|}{ScaleFL \cite{scalefl}}                                                                                   & \multicolumn{3}{c|}{DR-FL (Ours)}                                                                                                 \\ \hline
Distribution & \multicolumn{1}{c|}{$\alpha$=0.1} & \multicolumn{1}{c|}{$\alpha$=0.5}          & \multicolumn{1}{c|}{$\alpha$=1.0}    & \multicolumn{1}{c|}{$\alpha$=0.1} & \multicolumn{1}{c|}{$\alpha$=0.5} & \multicolumn{1}{c|}{$\alpha$=1.0}  & \multicolumn{1}{c|}{$\alpha$=0.1}          & \multicolumn{1}{c|}{$\alpha$=0.5}          & \multicolumn{1}{c|}{$\alpha$=1.0}                    \\ \hline
Model\_1      & \multicolumn{1}{c|}{45.06 $\pm$ 2.01} & \multicolumn{1}{c|}{85.58 $\pm$ 1.31}          & \multicolumn{1}{c|}{87.00 $\pm$ 1.93}  & \multicolumn{1}{c|}{53.78 $\pm$ 0.98} & \multicolumn{1}{c|}{74.26 $\pm$ 2.34} & \multicolumn{1}{c|}{\textbf{87.29 $\pm$ 0.93}}     & \multicolumn{1}{c|}{\textbf{80.15 $\pm$ 0.23}} & \multicolumn{1}{c|}{\textbf{82.25 $\pm$ 0.19}} & \multicolumn{1}{c|}{87.10 $\pm$ 0.37} \\ \hline
Model\_2      & \multicolumn{1}{c|}{59.76 $\pm$ 0.46} & \multicolumn{1}{c|}{85.75 $\pm$ 0.63}          & \multicolumn{1}{c|}{88.60 $\pm$ 0.34}  & \multicolumn{1}{c|}{57.19 $\pm$ 3.13} & \multicolumn{1}{c|}{85.32 $\pm$ 2.51} & \multicolumn{1}{c|}{\textbf{87.44 $\pm$ 0.55}}     & \multicolumn{1}{c|}{\textbf{82.10 $\pm$ 0.39}} & \multicolumn{1}{c|}{\textbf{88.76 $\pm$ 0.23}} & \multicolumn{1}{c|}{85.22 $\pm$ 0.34} 
\\ \hline
Model\_3      & \multicolumn{1}{c|}{57.25 $\pm$ 0.98} & \multicolumn{1}{c|}{83.26 $\pm$ 3.27}          & \multicolumn{1}{c|}{87.75 $\pm$ 1.25} & \multicolumn{1}{c|}{62.26 $\pm$ 1.34} & \multicolumn{1}{c|}{87.69 $\pm$ 1.07} & \multicolumn{1}{c|}{88.47 $\pm$ 0.97}    & \multicolumn{1}{c|}{\textbf{86.88 $\pm$ 0.23}} & \multicolumn{1}{c|}{\textbf{89.34 $\pm$ 0.62}} & \multicolumn{1}{c|}{\textbf{90.52 $\pm$ 0.13}} \\ \hline
Model\_4      & \multicolumn{1}{c|}{56.32 $\pm$ 4.07} & \multicolumn{1}{c|}{87.82 $\pm$ 1.28}          & \multicolumn{1}{c|}{87.83 $\pm$ 0.56} & \multicolumn{1}{c|}{55.85 $\pm$ 1.51} & \multicolumn{1}{c|}{86.78 $\pm$ 3.27} & \multicolumn{1}{c|}{88.40 $\pm$ 0.69}     & \multicolumn{1}{c|}{\textbf{85.80 $\pm$ 0.17}} & \multicolumn{1}{c|}{\textbf{89.36 $\pm$ 0.11}} & \multicolumn{1}{c|}{\textbf{89.60 $\pm$ 0.29}} \\ \hline
\end{tabular}
\label{test_acc}
\vspace{-0.10in}
\end{table*}

\section{Experimental Results}
\label{exp}
In order to evaluate the effectiveness of the approach we propose, we utilized the DR-FL algorithm by employing PyTorch with version 1.4.0.
 Like FedAvg, we make the assumption that only 10\% of AIoT devices participated in each round of FL communication throughout the training period.
 Regarding the case of DR-FL and other heterogeneous FL algorithms, we assign a small batch size of 32. 
 The local training epochs and initial learning rate were set at 5 and 0.05, respectively.
 In order to model a range of energy-limited situations, we assume that every gadget is equipped with a battery with a maximum capacity of 7,560 joules.
 To be precise, the capacity of each battery is 1500 mA with a rated voltage of 5.04V. 
 We conducted extensive experiments to address the following four Research Questions (RQs). 

\textbf{RQ1: (Superiority of DR-FL)}: What advantages can DR-FL achieve compared to state-of-the-art heterogeneous FL methods?
 
\textbf{RQ2: (Advantages of  MARL-based Dual-Selection?)} What advantages does MARL-based Dual-Selection offer in DR-FL procedure, particularly when dealing with limitations such as device energy and total training time, in comparison to other state-of-the-art heterogeneous FL methods?

\textbf{RQ3: (Scalability of DR-FL)}: What is the impact of the quantity of AIoT devices engaged in knowledge sharing on the performance of DR-FL?

\textbf{RQ4: (Investigation of the Validation Data Ratio)}: What is the impact of varying the proportion of validation data in MARL on the performance of DR-FL?

\subsection{Experimental Settings}
\subsubsection{Model Settings}
We conducted a comparison between our DR-FL approach and two well-known state-of-the-art heterogeneous FL methods, namely HeteroFL \cite{iclr_diao2021} and ScaleFL \cite{scalefl}. HeteroFL falls under the category of subnetwork aggregation-based methods, while ScaleFL belongs to the knowledge distillation-based methods. 
The ResNet-18 model \cite{ResNet} serves as the backbone. Each block of the ResNet-18 model is accompanied by a bottleneck and classifier, resulting in the creation of four distinct layer-wise models. These models are designed to simulate four different types of heterogeneous models, referred to as Models 1-4 in Table~\ref{test_acc}. 
Note that each layer-wise model can be reused with the same backbone for the purpose of model inference.

\subsubsection{Dataset Settings}
To evaluate
the effectiveness of  DR-FL,  we considered four training datasets: i.e., CIFAR10, CIFAR100 \cite{CIFAR}, Street View House Numbers (SVHN) \cite{SVHN}, Fashion-MNIST \cite{Fashion-mnist}.  CIFAR10: The CIFAR10 dataset consists of 60,000 $32\times32$ colour images across ten classes, with 6,000 images per class. The dataset is split into 50,000 training images and 10,000 testing images. 
CIFAR100: The CIFAR100 dataset is similar to CIFAR10 but contains 100 classes instead of 10, with 600 images per class. The dataset also comprises 50,000 training images and 10,000 testing images. 
SVHN: The SVHN dataset is a real-world image dataset derived from house numbers in Google Street View images. It contains over 600,000 labelled digit images, where each image is a 32×32 colour image representing a single digit (0-9).
Fashion-MNIST: The Fashion-MNIST dataset is a dataset of digital number images \cite{Fashion-mnist}, consisting of 70,000 $28\times28$ grayscale images of 10 different fashion categories. 
In our subsequent assessments, we evaluated three non-Independent and Identically Distributed (non-IID) distributions for each dataset. 
Following the methodology of HeteroFL as described in \cite{iclr_diao2021}, we generated non-IID local training datasets by utilizing heterogeneous data splits. These splits were created based on a Dirichlet distribution, which was controlled by a variable $\alpha$. A smaller value of $\alpha$ often indicates a greater level of non-IID distribution.
Additionally, we employed the same data augmentation techniques as those utilized in HeteroFL \cite{iclr_diao2021} to maximize the efficient use of natural picture datasets. 
To facilitate Multi-Agent Reinforcement Learning (MARL) training on a cloud server in DR-FL, we allocated 4\% of the total training data as the validation set on the server. It is important to understand that the validation set used on the server is completely separate from the local training datasets stored on AIoT devices. 

\subsubsection{Test-bed Settings} 
\begin{figure}[H]
    \centering
    \begin{subfigure}{0.49\linewidth}
        \centering
        \includegraphics[width=\linewidth]{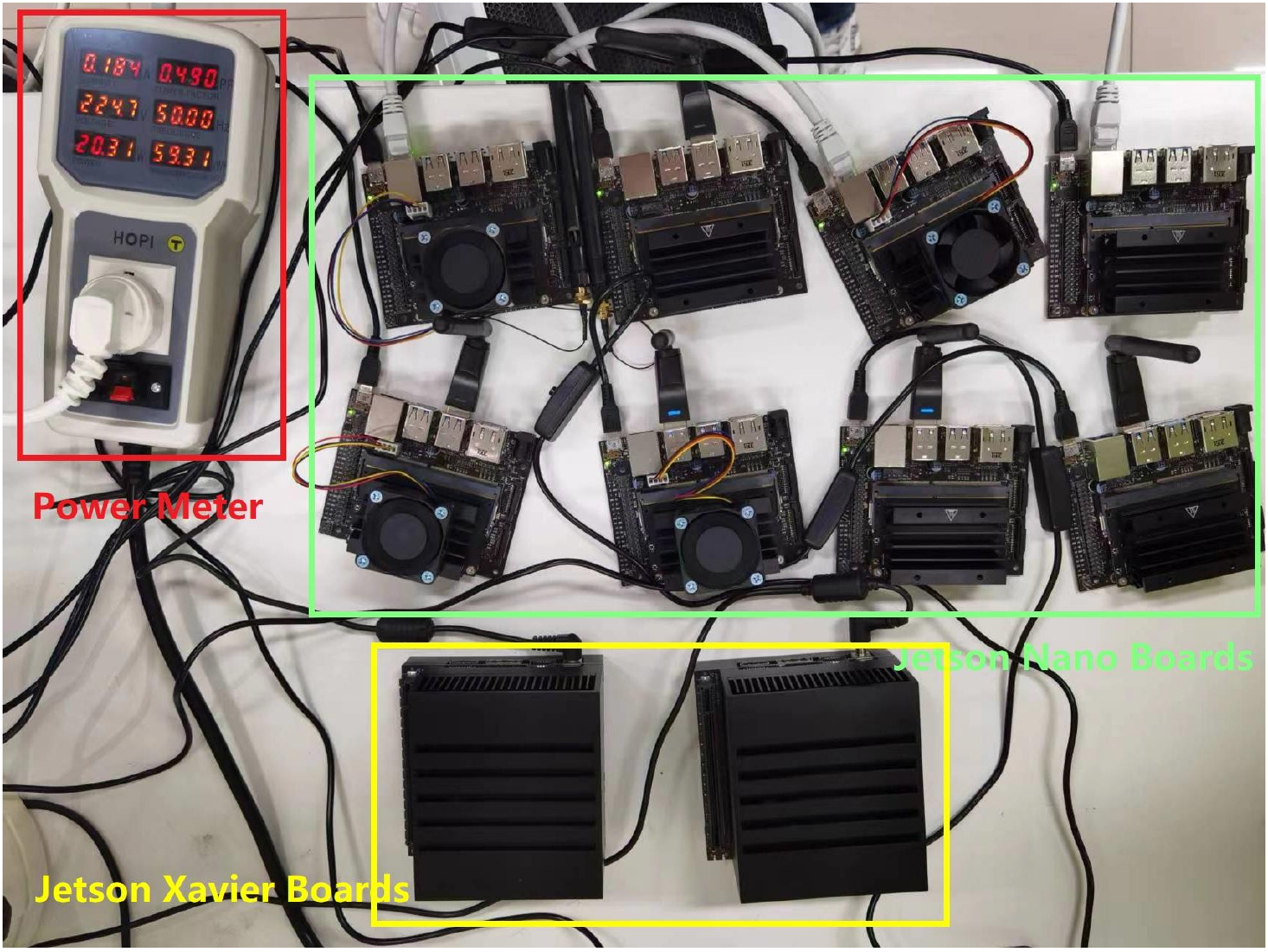}
        \caption{AIoT devices}
    \end{subfigure}
    \hfill
    \begin{subfigure}{0.49\linewidth}
        \centering
        \includegraphics[width=\linewidth]{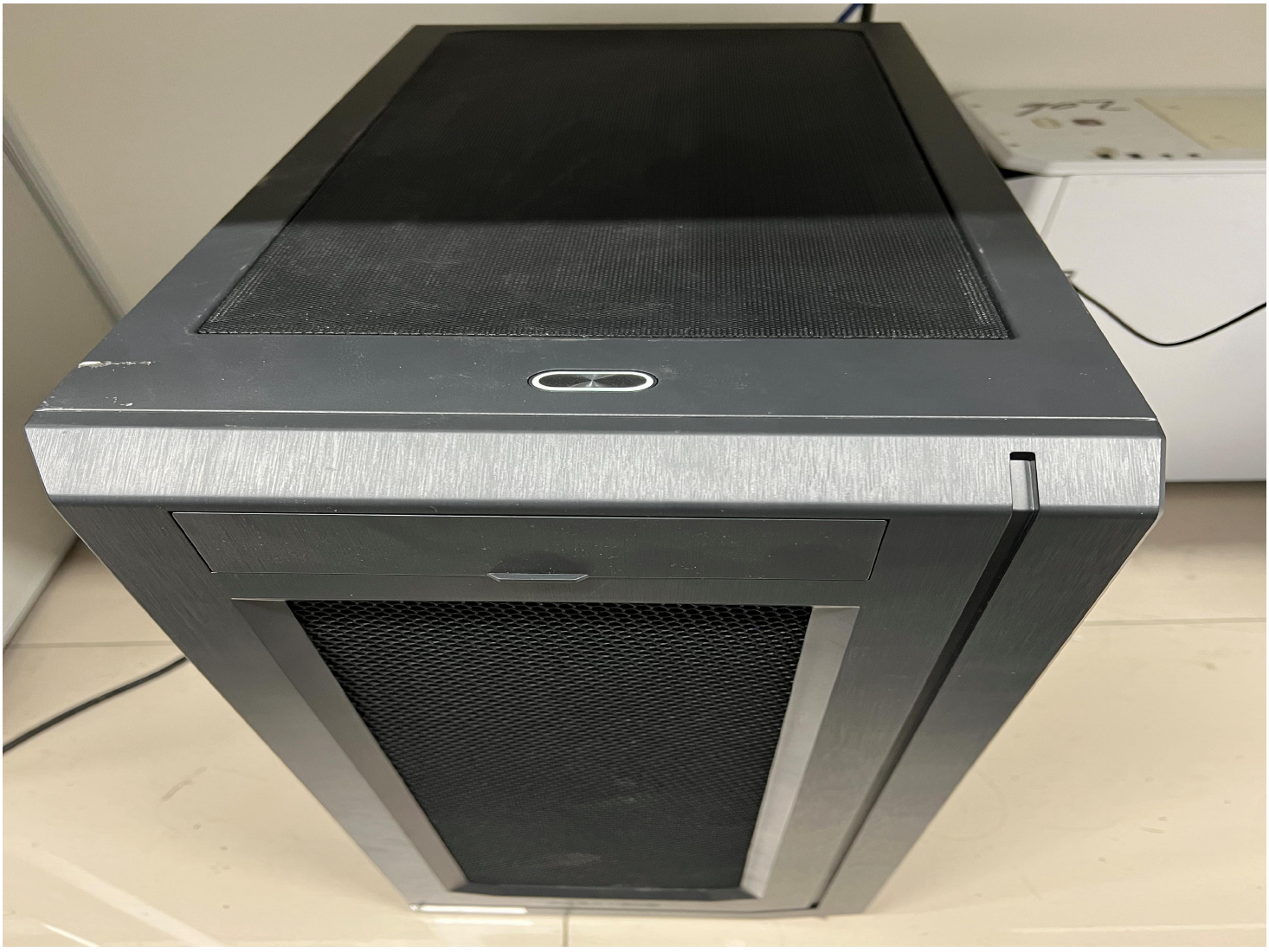}
        \caption{The server}
    \end{subfigure}
    \vspace{-0.15in}
    \caption{Real test-bed platform for our experiment.}
    \vspace{-0.2in}
    \label{fig:real-environment}
\end{figure}

In addition to conducting simulation-based evaluation, we developed a physical test-bed platform, depicted in Figure \ref{fig:real-environment}, to assess the effectiveness of our DR-FL in a real-world setting. 
The test-bed platform comprises four components:  i) The cloud server is constructed on an Ubuntu workstation that has an Intel i9 CPU, 32G memory, and a GTX3090 GPU. ii) The Jetson Nano boards each contain a quad-core ARM A57 CPU, a 128-core NVIDIA Maxwell GPU, and 4GB LPDDR4 RAM. iii) The Jetson AGX Xavier boards are equipped with an 8-core CPU and a 512-core Volta GPU. iv) Shenzhen HOPI Electronic Technology Ltd. produces the HP 9800 power meter, which is located in the top-left part of Figure \ref{fig:real-environment}(a). It is important to mention that, in addition to the federated training process, we utilized a power meter to accurately measure the energy consumption of all the AIoT devices at one-second intervals during the development of the MARL environment. 

\subsection{Accuracy Comparison (RQ1)}
To evaluate the effectiveness of our proposed DR-FL, Table \ref{test_acc_energy} 
presents the best test accuracy information for HeteroFL, ScaleFL and our DR-FL under the specific energy constraints along the FL processes based on the four datasets, assuming all the device batteries are initialized to be full.
  For each dataset and FL method combination, we considered three kinds of data distributions for all local AIoT devices, where the non-IID settings follow the Dirichlet distributions controlled by  $\alpha$. 
Note that the baseline approaches (HeteroFL and ScaleFL) do not consider the energetic constraints in their FL procedure. 
To make a fair comparison, we added the greedy algorithm for energy awareness in this experiment (model selection will select the maximum model that can be trained in FL) into the two baseline algorithms for comparison. The experiments were repeated five times to calculate the mean and variance. 

From Table \ref{test_acc_energy}, it is evident that within the constraint of the restricted battery energy conditions set for each device, DR-FL  exhibits superior inference performance, surpassing results in 29 out of the 36 evaluated scenarios in comparison with other baseline algorithms. 
Specifically,  no matter which data set, in the scenario of $\alpha=0.1$, our method shows superior performance in comparison with other baseline algorithms. 
Moreover, the performance of some models at $\alpha = 0.1$   in DR-FL has exceeded the performance of two baselines  at $\alpha = 0.5$.
As an example shown in the non-IID scenario of SVHN with  $\alpha=0.1$, the test accuracy of DR-FL reaches 91.47\%, while HeteroFL only attains 66.31\% and ScaleFL only gets 76.73\% on Model\_3.
This is  because
our MARL-based dual-selection method can efficiently utilize the available energy of devices by 
assigning specific layer-wise models to participating devices that are more suitable for heterogeneous federated learning.

\subsection{Comparison of Energy  Consumption  (RQ2)}

\begin{figure}[h]
    \vspace{-0.05in}
    \centering
    \begin{subfigure}[b]{0.491\linewidth}
        \centering
        \includegraphics[width=\linewidth]{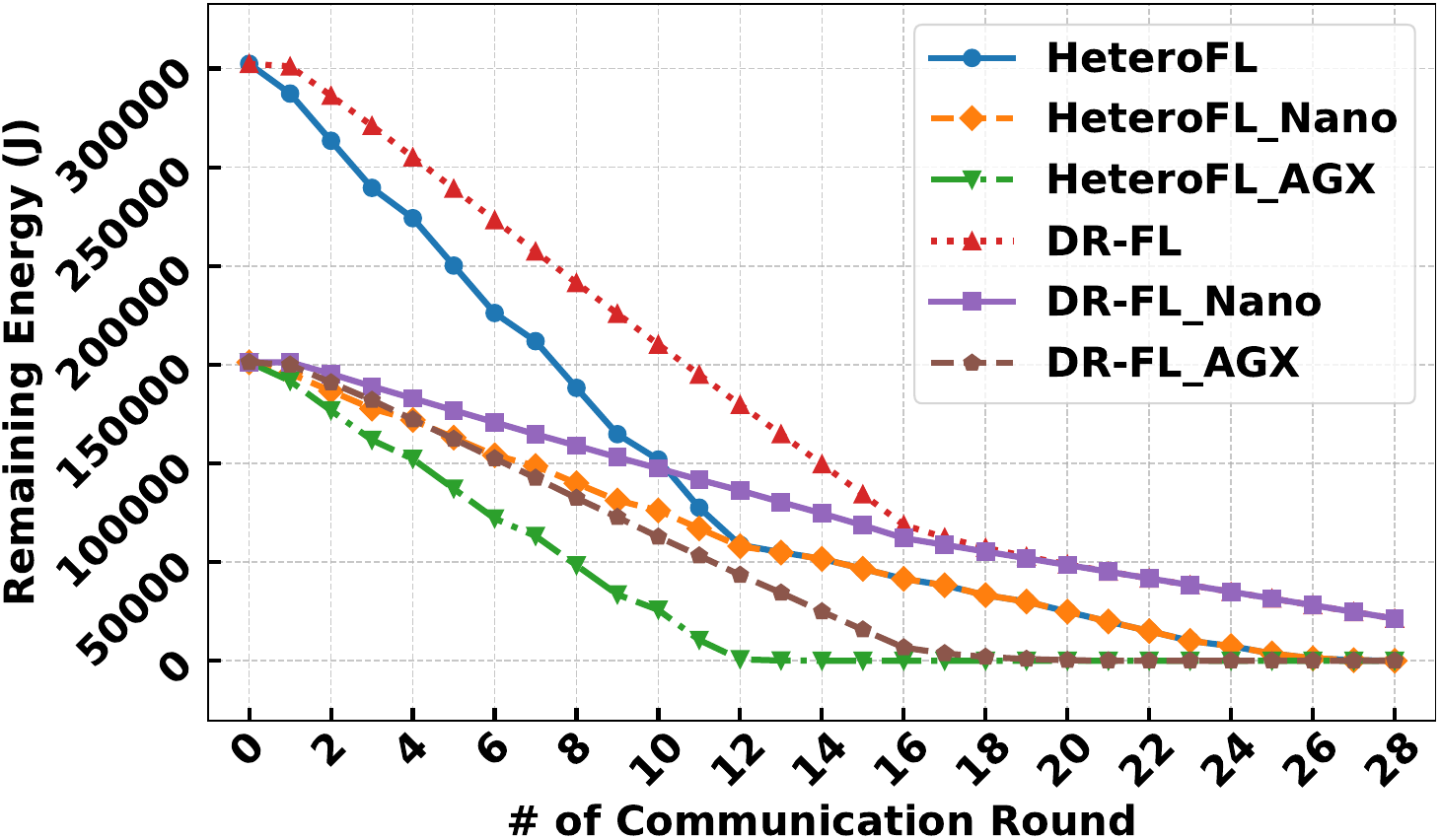}
        \caption{\scriptsize Total energy variation}
    \end{subfigure}
    \hfill
    \begin{subfigure}[b]{0.491\linewidth}
        \centering
        \includegraphics[width=\linewidth]{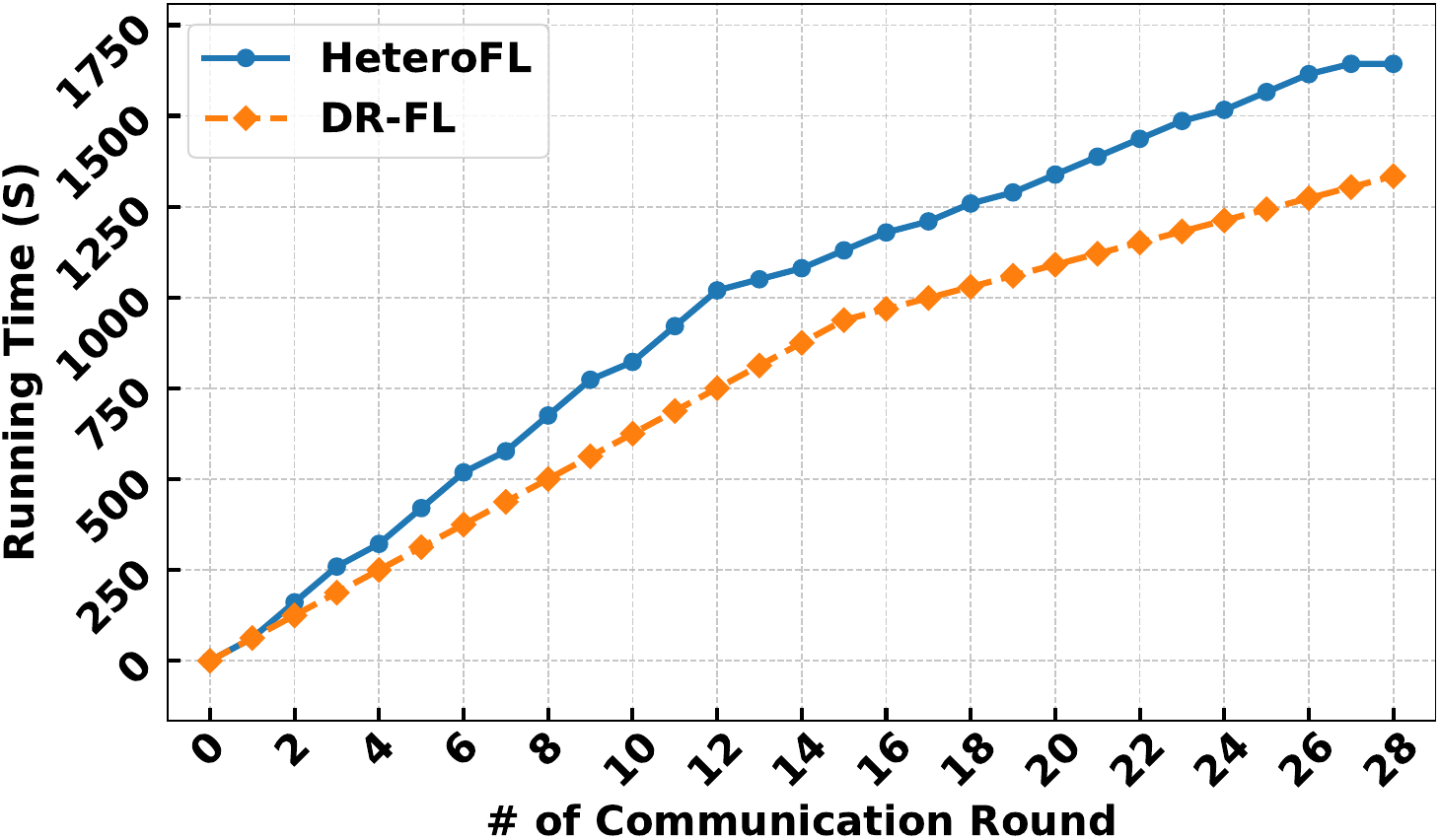}
        \caption{\scriptsize Total running time}
    \end{subfigure}
    \vspace{-0.1in}
    \caption{Comparison of total energy consumption and running time.}
    \label{comparison_energy}
    \vspace{-0.1in}
\end{figure}

To evaluate the effectiveness of our DR-FL technique in terms of energy usage and execution time, we carried out an experiment comprising a total of 40 devices, specifically 20 Jetson Nano boards and 20 AGX Xavier boards. 
Figure \ref{comparison_energy} illustrates the differences in total remaining energy variation and running time between the federated learning processes employing HeteroFL (where ScaleFL has the same energy consumption and running time as the greedy algorithm) and DR-FL. 
Each subfigure is represented by the notation $X\_Y$, which denotes the cumulative outcome of all devices of type $Y$ employing technique $X$.
If the variable $Y$ is excluded, the term $X$ represents the overall outcome that encompasses all the devices.
For instance, in Figure \ref{comparison_energy}(a), the label \textit{DR-FL} indicates the total remaining energy of the 40 devices, whereas \textit{DR-FL\_Nano} represents the total remaining energy of the 20 Jetson Nano boards.

Figure \ref{comparison_energy}(a) demonstrates that our approach can accommodate a greater number of training rounds while adhering to the same energy limitations. Consequently, this results in improved test accuracy and energy efficiency. 
As an illustration, in the case of HeteroFL, the devices powered by Jetson AGX Xavier became depleted of battery power by the 12th round. Nevertheless, during the DR-FL event, the devices powered by Jetson AGX Xavier experienced battery depletion by the 18th round.
Furthermore, in Figure \ref{comparison_energy}(b), there is a distinct inflexion point observed in the 12$^{th}$ round for HeteroFL. Subsequently, only devices based on Jetson Nano are utilized for federated training. 
Nevertheless, in the case of DR-FL, a notable turning point can be observed in the 15$^{th}$ cycle, which signifies the efficacy of the MARL algorithm in managing the energy wastage of the device by minimizing useless waiting and training time.

\subsection{Scalability Analysis (RQ3)}

\begin{figure}[htbp]
\centering
    \subfloat[\scriptsize{CIFAR10 ($\alpha=0.1$) w/ 40 devices}]{
        \includegraphics[width=0.495\linewidth]{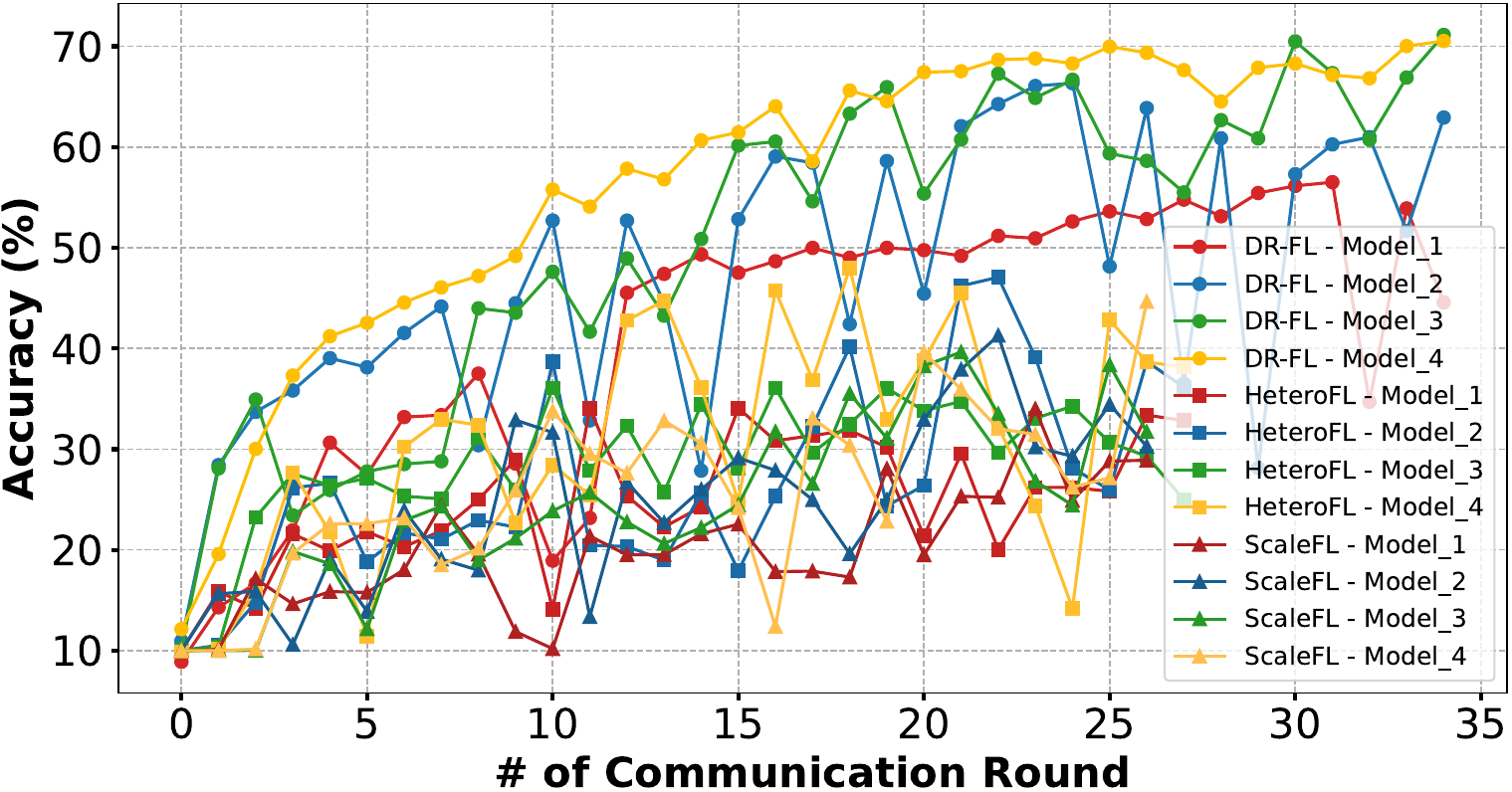}}
    \subfloat[\scriptsize{CIFAR10 ($\alpha=0.1$) w/ 60 devices}]{
        \includegraphics[width=0.495\linewidth]{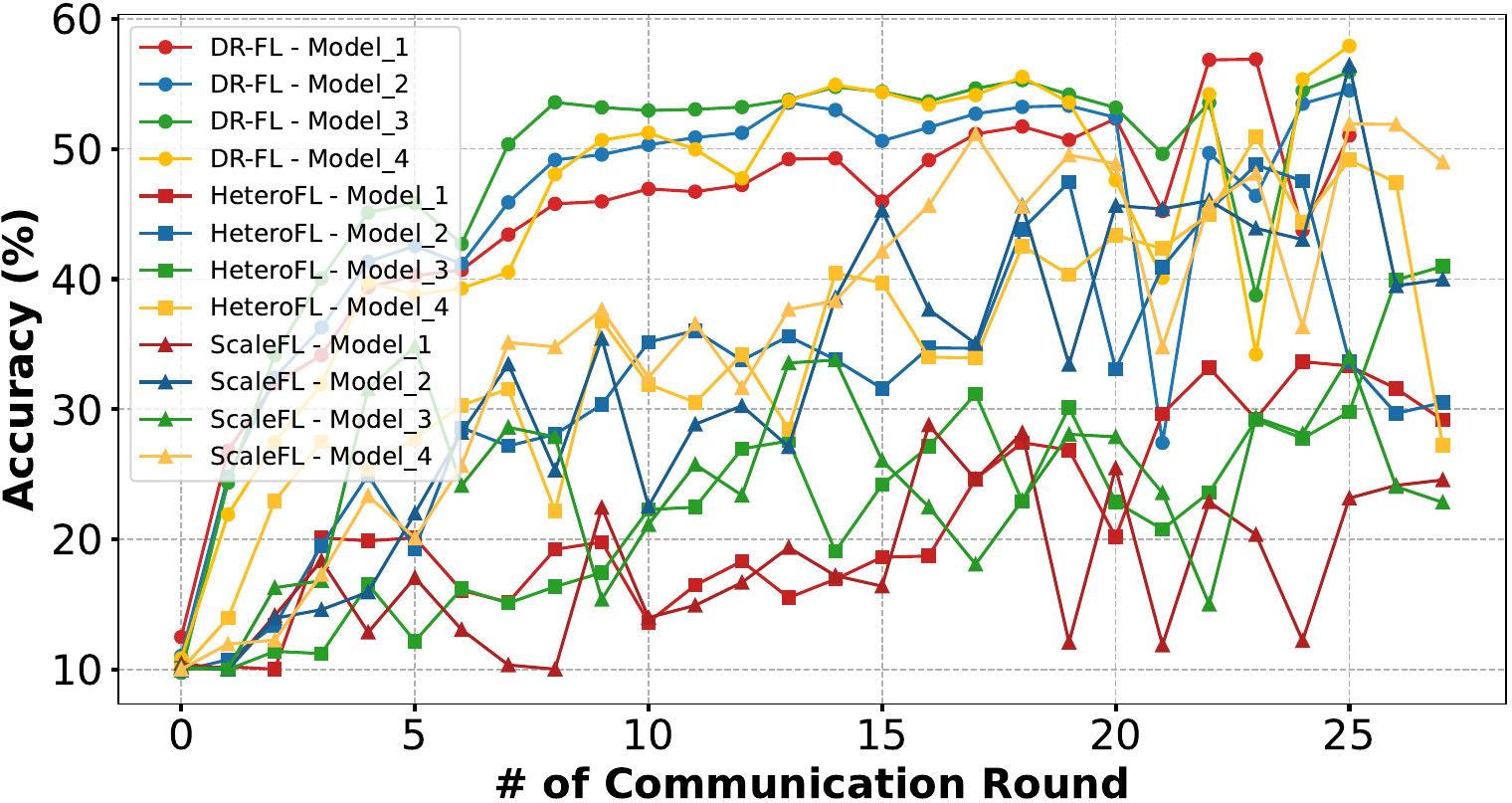}}

       \subfloat[\scriptsize{Fashion-MNIST ($\alpha=0.1$) w/ 40 devices}]{
        \includegraphics[width=0.495\linewidth]{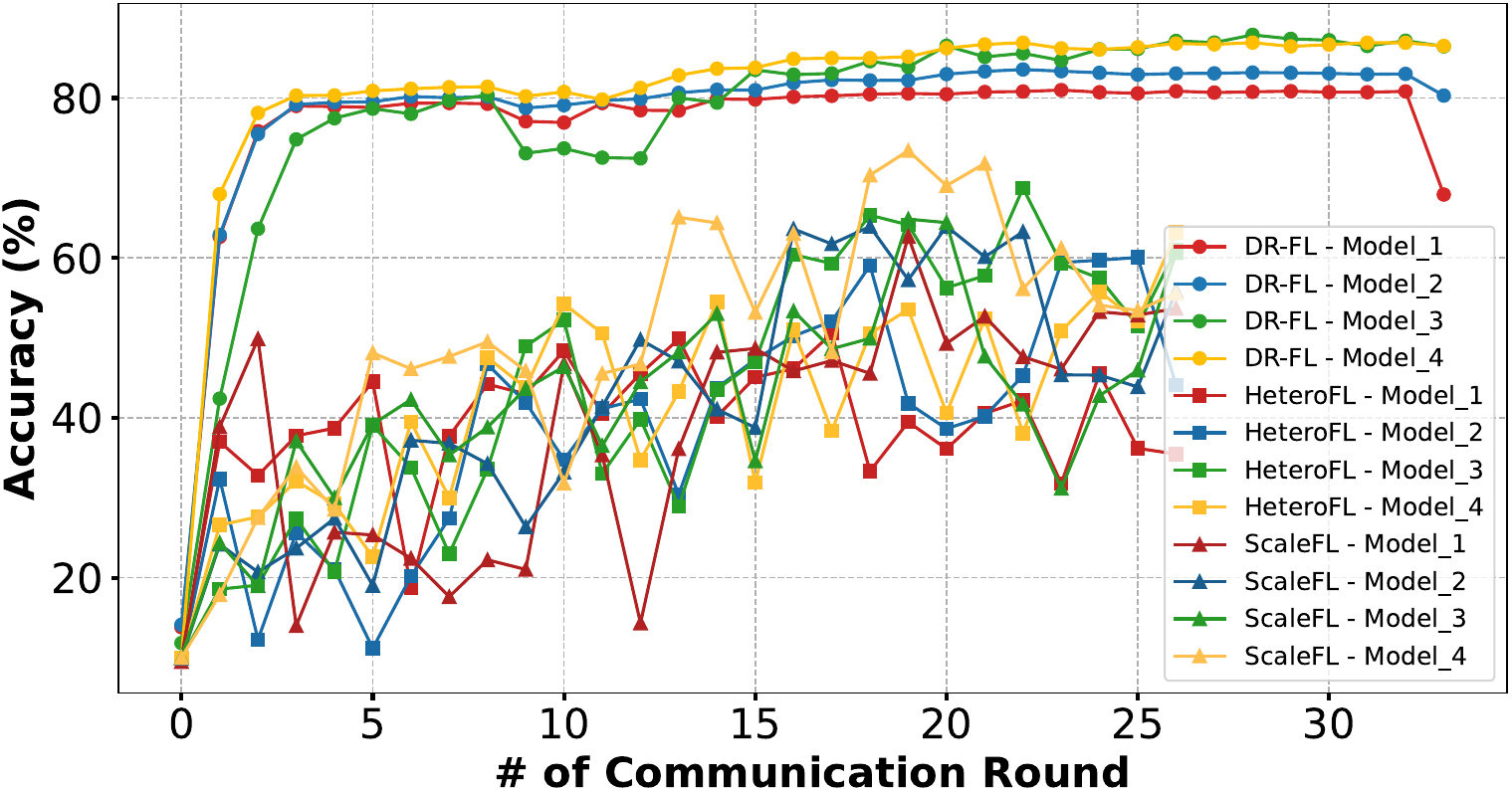}}
    \subfloat[\scriptsize{Fashion-MNIST ($\alpha=0.1$) w/ 60 devices}]{
        \includegraphics[width=0.495\linewidth]{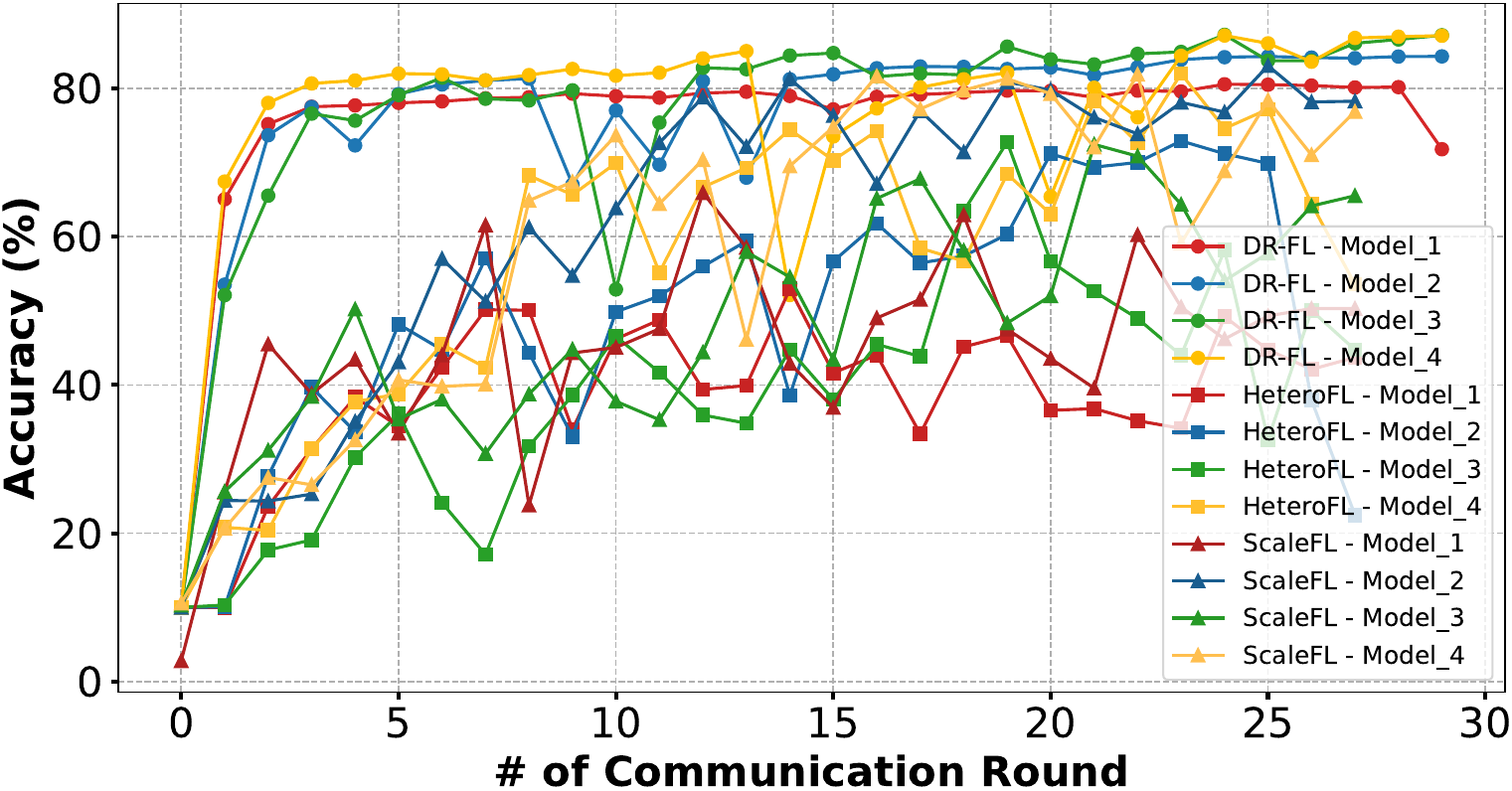}}
        
        \subfloat[\scriptsize{SVHN ($\alpha=0.1$) w/ 40 devices}]{
        \includegraphics[width=0.495\linewidth]{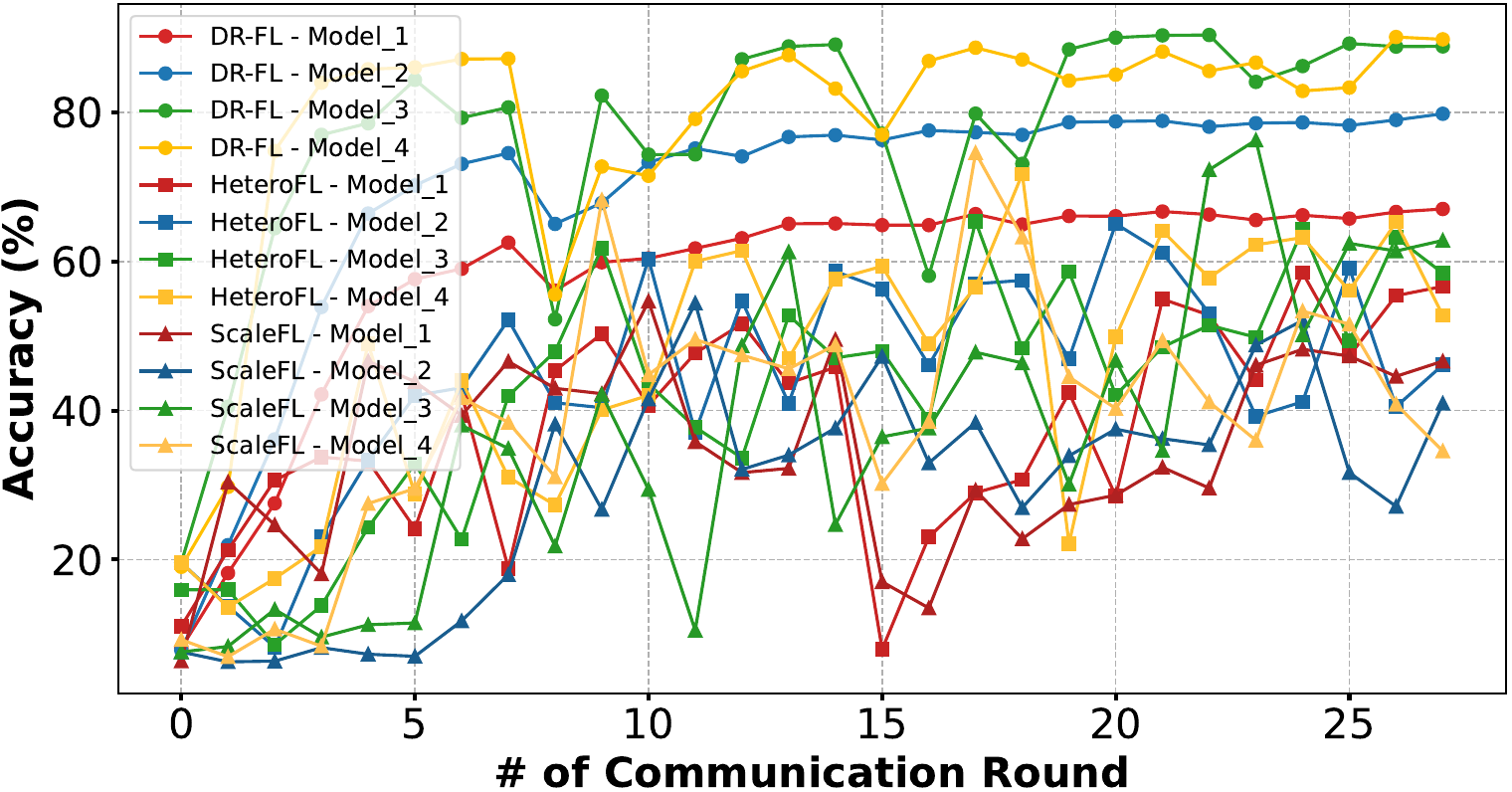}}
    \subfloat[\scriptsize{SVHN ($\alpha=0.1$) w/ 60 devices}]{
        \includegraphics[width=0.495\linewidth]{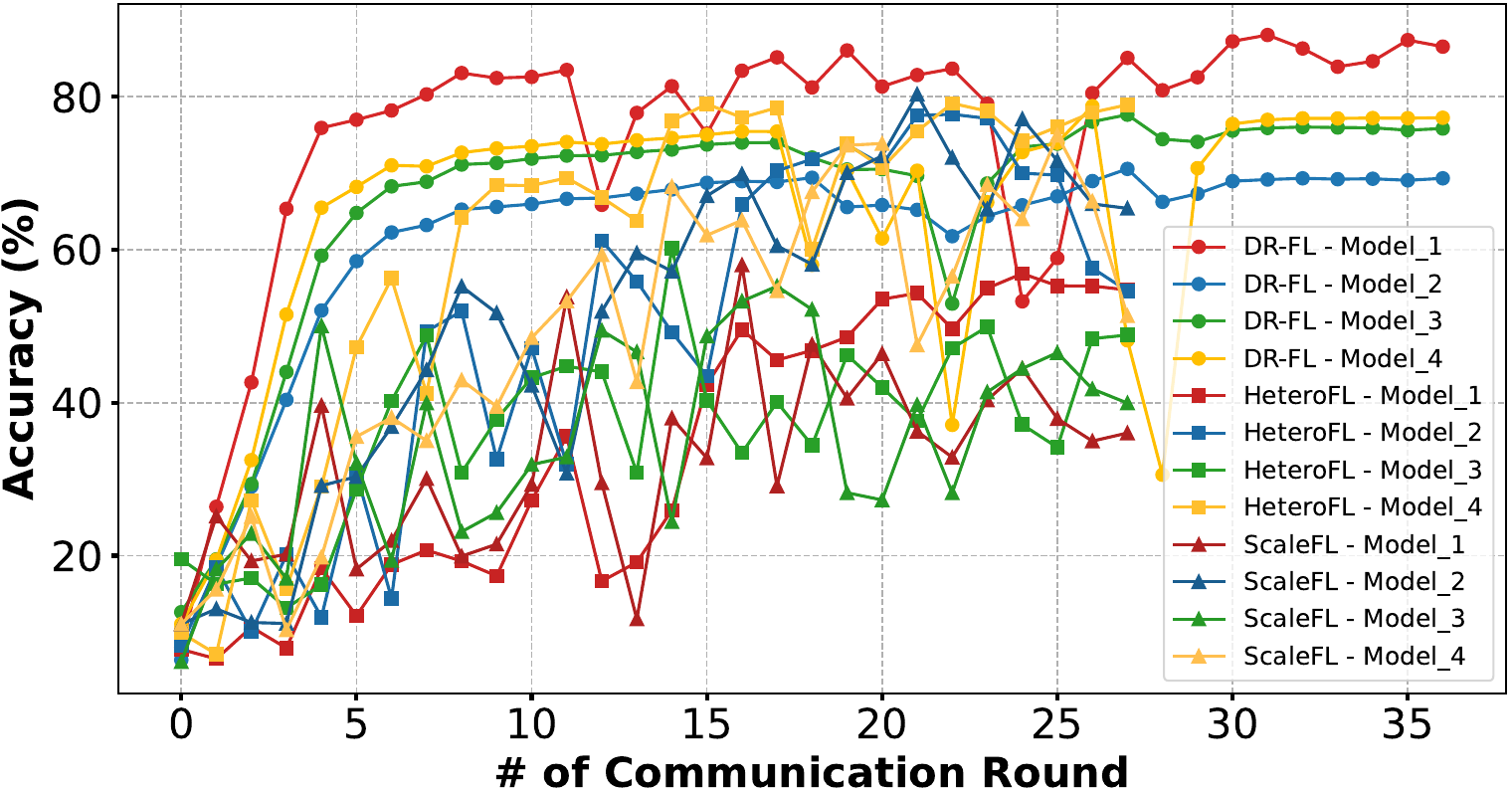}}
\vspace{-0.15in}
\caption{Learning curves of DR-FL and other baselines in AIoT systems with different numbers of devices under limited energy constraints.}
\label{sca_result} 
\vspace{-0.2in}
\end{figure}

Figure ~\ref{sca_result} illustrates the test accuracy of three approaches (HeteroFL, ScaleFL, and DR-FL) in various non-IID scenarios with different numbers of devices, all within specified energy constraints. 
From this figure, we can observe that when more heterogeneous devices participate in FL, the superiority of DR-FL becomes more 
significant than that of the other two methods.
For example, for the non-IID scenario of CIFAR10, Fashion-MNIST and SVHN (with $\alpha$=0.1), DR-FL consistently achieves higher test accuracy than ScaleFL and HeteroFL.

\subsection{Ablation Study  (RQ4)}

To explore the role of the validation set proportion in our method, the validation set with different proportions (1\%-10\%) is selected for the experiment of this paper, and the non-independent data set CIFAR10 ($\alpha=0.1$) is selected as the exploration scenario. From Table~\ref{tab:accuracy}, we can see, with the number of validation set increases, in the initial overall test accuracy rise, and with the proportion of validation sets more than 4\%, the accuracy decreases. 
This phenomenon shows that it can be used as an effective tuning knob to explore the trade-off between the proportion of cloud validation data and the entire DR-FL performance. We found that the setup validation data ratio of 4\% provided a reasonable balance. We picked 4\% and used it in all experiments.

\begin{table}[h!]
 \vspace{-0.15in}
    \centering
    \caption{Average model accuracy with different percentages of the validation dataset}
    \vspace{-0.1in}
    \resizebox{0.48\textwidth}{!}{
    \begin{tabular}{|c|c|c|c|c|c|c|c|c|c|c|}
        \hline
        \textbf{Percentage} & 1\% & 2\% & 3\% & 4\% & 5\% & 6\% & 7\% & 8\% & 9\% & 10\% \\
        \hline
        \textbf{Accuracy (acc)} & 57.72 & 63.23 & 64.35 & \textbf{65.04} & 63.16 & 59.18 & 58.86 & 52.21 & 54.99 & 55.69 \\
        \hline
    \end{tabular}
}
    \label{tab:accuracy}
    \vspace{-0.2in}
\end{table}

\section{Conclusion}
\label{con}

Federated Learning (FL) is intended to facilitate privacy-preserving collaborative learning among Artificial Intelligence of Things (AIoT) devices.  Nevertheless, the current design of AIoT systems based on Federated Learning (FL) encounters significant challenges, such as non-IID data, heterogeneous local devices, and varying computational and energy capabilities. Consequently, these challenges result in issues such as low inference accuracy, excessive battery consumption, and increased training time. 
 This paper presents an innovative Federated Learning (FL) paradigm that facilitates effective information exchange among various devices while considering unique energy limitations.
Our proposed layer-wise aggregation method and MARL-based dual selection mechanism enable AIoT devices with varying computational and energy capabilities to intelligently choose suitable local models for global model training. This allows devices to effectively learn from each other by utilizing relevant components from different layer-wise models. 
 The efficacy of DR-FL in terms of inference performance, energy consumption, and scalability has been demonstrated through extensive experiments conducted on widely recognized datasets.


\bibliographystyle{ACM-Reference-Format}
\bibliography{sample-base}



\end{document}